\documentclass[10pt,twocolumn,letterpaper]{article}

\usepackage{stfloats, multicol}
\usepackage{lipsum}

\usepackage{booktabs} 
\usepackage{makecell}
\usepackage{cellspace} 
\usepackage{pifont}
\newcolumntype{P}[1]{>{\cellspacetoplimit=10pt\cellspacebottomlimit=4pt}p{#1}}

\newcommand{\xmark}{\ding{55}}%

\usepackage{multirow}

\usepackage{subcaption}

\usepackage[pagenumbers]{cvpr} 

\usepackage[dvipsnames]{xcolor}

\definecolor{cvprblue}{rgb}{0.21,0.49,0.74}
\usepackage[pagebackref,breaklinks,colorlinks,citecolor=cvprblue]{hyperref}

\def\paperID{6685} 
\def\confName{CVPR}
\def\confYear{2024}

\title{Action Scene Graphs for Long-Form Understanding of Egocentric Videos}

\author{Ivan Rodin$^{\ast\,1}$ \quad Antonino Furnari$^{\ast\,1}$ \quad Kyle Min$^{\ast\,2}$ \quad Subarna Tripathi$^2$ \quad Giovanni Maria Farinella$^1$ \\[1ex]
$^1$University of Catania \qquad $^2$Intel Labs\\
{\tt\footnotesize \{ivan.rodin,antonino.furnari,giovanni.farinella\}@unict.it} \quad \tt\footnotesize \{kyle.min,subarna.tripathi\}@intel.com
}

\begin{document}

\twocolumn[{%
\renewcommand\twocolumn[1][]{#1}%
\maketitle
\begin{center}
\vspace{-0.8cm}
    \centering
    \captionsetup{type=figure}
    \includegraphics[width=0.96\textwidth]{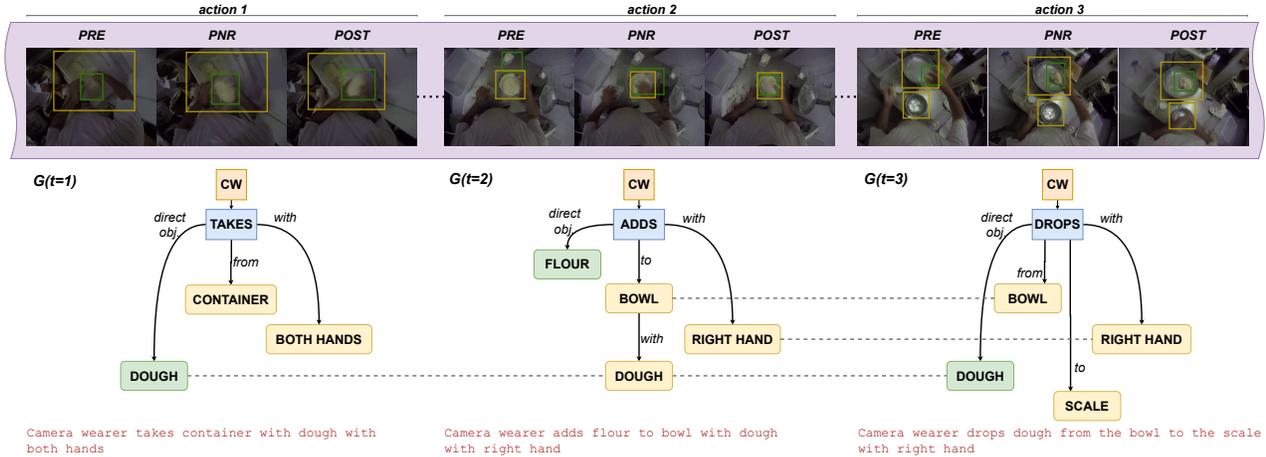}
    \captionof{figure}{Egocentric Action Scene Graphs are temporal dynamic graphs ($G(t)$) capturing the action verbs (nodes in blue), direct or active objects
(nodes in green), and other objects (nodes in yellow) involved in the activity performed by a camera wearer (the orange CW node).
Edges between nodes represent relationship between the verb and the objects or between object pairs.
The graph evolves through time 
providing a long-from representation of the egocentric video (dashed lines). 
Objects of interaction are grounded
with bounding boxes.
}
\label{fig:easg}
\end{center}%
}]
\let\thefootnote\relax\footnote{$^{\ast}$These authors contributed equally to this work.}
\begin{abstract}
We present Egocentric Action Scene Graphs (EASGs), a new representation for long-form understanding of egocentric videos. EASGs extend standard manually-annotated representations of egocentric videos, such as verb-noun action labels, by providing a temporally evolving graph-based description of the actions performed by the camera wearer, including interacted objects, their relationships, and how actions unfold in time. Through a novel annotation procedure, we extend the Ego4D dataset by adding manually labeled Egocentric Action Scene Graphs offering a rich set of annotations designed for long-from egocentric video understanding. We hence define the EASG generation task and provide a baseline approach, establishing preliminary benchmarks. Experiments on two downstream tasks, egocentric action anticipation and egocentric activity summarization, highlight the effectiveness of EASGs for long-form egocentric video understanding. We will release the dataset and the code to replicate experiments and annotations$^1$\footnote{$^{1}$GitHub page: \href{https://github.com/fpv-iplab/EASG}{https://github.com/fpv-iplab/EASG}}.


\end{abstract}
\section{Introduction}
\label{sec:intro}

Wearable devices allow to capture video of human activities from an egocentric perspective. A proper analysis of such video can enable a detailed understanding of how humans interact with the environment, how they manipulate objects, and, ultimately, what are their goals and intentions. Easily covering sequences of activities performed by the camera wearer in different physical locations, egocentric video is by its own nature \textit{long-form}~\cite{wu2021towards}. Hence, typical applications of egocentric vision systems require algorithms able to represent and process video over temporal spans that last in the order of minutes or hours. Examples of such applications are action anticipation~\cite{damen2018scaling,grauman2022ego4d,rodin2021predicting}, video summarization~\cite{del2016summarization}, and episodic memory retrieval~\cite{grauman2022ego4d}.
Despite the relevance of such applications in the panorama of egocentric vision~\cite{plizzari2023outlook}, progress in this area has been hindered by the lack of a comprehensive and long-form representation of videos that algorithms can rely on, with popular high-level human-gathered representations being in the form of textual narrations~\cite{damen2018scaling}, verb-noun action labels~\cite{fathi2011learning}, temporal bounds for action segments~\cite{fathi2011learning,li2018eye,damen2018scaling}, object bounding boxes~\cite{pirsiavash2012detecting}, object state changes~\cite{grauman2022ego4d}, and hand-object interaction states~\cite{shan2020understanding,darkhalil2022epic}, all short-range representations describing temporal spans lasting few seconds.

In this paper, we introduce a novel graph-based representation of actions performed by the camera wearer in an egocentric video, which we term \textit{Egocentric Action Scene Graph (EASG)}. The proposed representation builds on the literature of scene graphs~\cite{johnson2015image,ji2020action,rai2021home} to extend the classic \textit{verb-noun} action representation available in egocentric vision datasets~\cite{fathi2011learning,li2018eye,damen2018scaling,damen2020rescaling,grauman2022ego4d} to a structured format in which a sequence of actions performed by the camera wearer is represented with a \textit{temporal dynamic graph} encoding and grounding to the video the objects involved in the action, the action verb, and the main relationships between the considered objects (see Figure~\ref{fig:easg}). EASGs naturally model the temporal evolution of egocentric actions, thus providing a rich representation to be exploited in a variety of tasks requiring long-form video understanding. 

We build on the Ego4D dataset \cite{grauman2022ego4d}, which provides egocentric videos of individuals engaged in a range of activities representative of human perception, and augment it with manually gathered egocentric action scene graph labels collected through a novel annotation procedure involving different labeling steps and a validation stage.
As customary in the scene graph literature~\cite{ji2020action,yang2022panoptic}, we benchmark the egocentric action scene graph generation task both to provide baseline results and as a means of investigating the feasibility of automatically recovering such rich human-annotated representations, a fundamental ability for downstream applications. We hence show initial results highlighting the effectiveness of the proposed EASG representation in tackling long-form video understanding tasks such as action anticipation and activity summarization.

The contributions of this paper are as follows: 1) We introduce Egocentric Action Scene Graphs, a novel representation for long-form understanding of egocentric videos; 2) We extend Ego4D with manually annotated EASG labels, which are gathered through a novel annotation procedure; 3) We propose a EASG generation baseline and provide initial baseline results; 4) We present experiments that highlight the effectiveness of the EASG representation for long-form egocentric video understanding. We will release the dataset and the code to replicate data annotation and the experiments (GitHub page: \href{https://github.com/fpv-iplab/EASG}{https://github.com/fpv-iplab/EASG}).
\section{Related works}\label{sec:related}

Our work is related to previous research lines which are revised in the following sections.

\subsection{Graph-based representations for video understanding}
The exploration of graph-based representations in image and video analysis has burgeoned over recent years, offering a structured approach to encapsulate complex relationships and interactions among element of the scene inherent in visual data. 
Seminal works~\cite{Wang_2018_ECCV,Yuan2017TemporalDG,johnson2018image, herzig2019canonical,wu2020adaptive} in this domain have focused on various methodologies to facilitate the transition from raw visual data to structured graph representations. For instance, 
graph structured data is leveraged for image synthesis in~\cite{johnson2018image, herzig2019canonical}, learning video representations in~\cite{Wang_2018_ECCV},  detecting video objects in~\cite{Yuan2017TemporalDG} and
person re-identification  in~\cite{wu2020adaptive}.
In \cite{tang2019learning}, the Visual Context Tree (VCTree) graph structure is built for the purpose of visual question answering. 
The work of \cite{lu2021context} utilizes a transformer architecture to produce a scene graph of an image, while \cite{cong2021spatial,li2022sgtr,feng2023exploiting,nag2023unbiased} explores approaches to build dynamic scene graphs to describe the scene based on the video inputs.

Scene graph generation extends beyond being an end goal, as a powerful precursor for downstream applications in computer vision, enabling enhanced performance in complex tasks. For instance, the works of~\cite{he2021exploiting} demonstrated how scene graph generation could be leveraged to improve object detection and visual relationship detection, thus offering a more contextual understanding of visual scenes. Similarly, the works of~\cite{nguyen2021defense,yang2019auto,yang2022reformer} explored the utilization of scene graphs for image captioning, where the generated graphs provided a structured semantic understanding that enriched the descriptive quality of generated captions.

The works of~\cite{mao2022dynamic,zhao2023constructing,arnab2021unified,ji2020action,rai2021home} delved into employing scene graphs for video understanding, showcasing that the structured representations facilitated a more nuanced understanding of temporal actions and interactions within videos. These collective efforts accentuate the instrumental role of scene graph generation not just as a standalone objective but as a potent enabler for a spectrum of downstream tasks, amplifying the scope and efficacy of visual understanding.

Building on previous investigations, in this work, we propose a novel graph-based representation for actions in egocentric videos. Our representation is shown to improve long-from video understanding in the considered domain.

\subsection{Graph-based representations in Egocentric Vision} Although there has been substantial work in video scene graph processing, only a limited number of studies have focused on egocentric videos. The unique perspective offered by egocentric videos allows for a different approach to understand human interactions and activities. 
Scene graphs from the perspective of autonomous vehicles have been more extensively researched by the community \cite{malawade2022roadscene2vec,kochakarn2023explainable} than the human-centric view videos. However, there are a few previous works studying the applicability of scene graph representation in egovision. 
In~\cite{min2022intel,min2023sthg}, the authors show how the graph-based representation can be useful for the audio-video diarization of egocentric videos.  
In \cite{lu2021multi}, the authors solve the problem of scene graph generation by composing exo- and ego-centric view processing to construct scene graphs. In \cite{singh2023scene}, egocentric scene graph representations are used to perform downstream tasks of embodied navigation. Ego-Topo \cite{nagarajan2020ego} presents graphs derived from egocentric videos, encoding the scene topology, thereby enhancing long-term video understanding and egocentric action anticipation. 
While these previous studies have shown the potential of extending scene graph generation techniques to egocentric videos, in this paper, we propose a general graph-based representation designed to be descriptive of human-object interactions happening in egocentric videos to improve long-form video understanding.

\begin{figure*}[htbp]
\centering
\includegraphics[width=\linewidth]{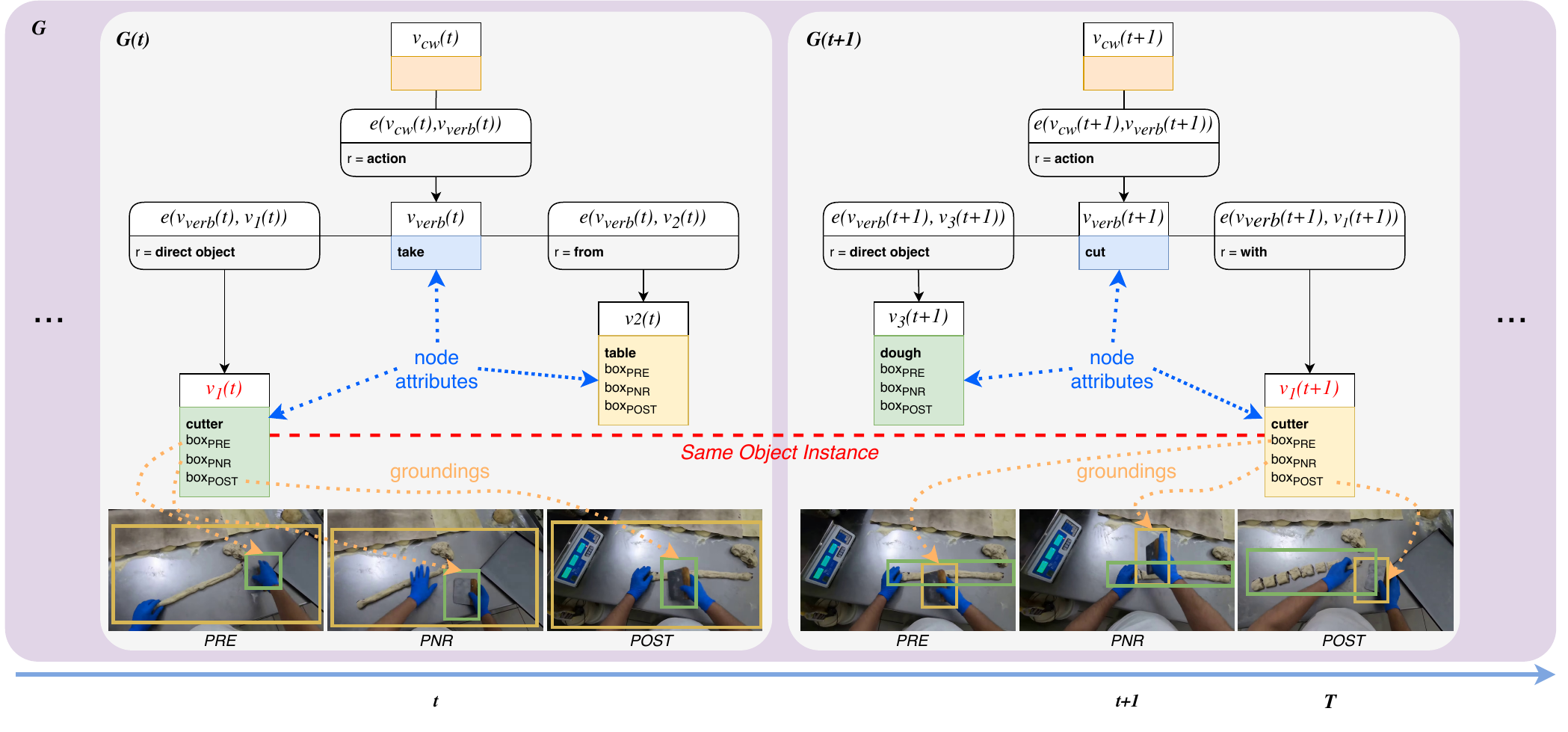} 
\caption{An Egocentric Action Scene Graph (EASG) is a time-varying directed graph $G(t)=(V(t),E(t))$, where nodes $V(t)$ represent either the camera wearer ($v_{cw}(t)$), the action verb ($v_{verb}(t)$), or the involved objects. Edges $E(t)$ represent relationships $e(v_i(t), v_j(t))$ between node pairs. Each node, except for the CW node, can have one or more attributes $att(v_j(t))$ (indicated in blue). Each object has three grounding bounding boxes in the $PRE$, $PNR$ and $POST$ frames (highlighted in orange). Nodes $v_j$ representing the same object instance maintain the same index across different timesteps (e.g., $v_1(t)$ and $v_1(t+1)$ highlighted in red).
}
\label{fig:notation}
\end{figure*}

\subsection{Graph-based Image and Video Datasets}
The introduction of image and video datasets containing graph-based scene representations has been instrumental to the development of graph-based representations.
Prominent datasets such as Action Genome \cite{ji2020action} and Home Action Genome \cite{rai2021home}, in particular, have contributed by providing rich graph structures that encode actions and interactions within videos. The Panoptic Scene Graph (PSG) dataset~\cite{yang2022panoptic} introduced enhanced annotations by replacing bounding boxes with fine-grained object segmentation masks. The Visual Genome dataset~\cite{krishna2017visual} has significantly contributed to the elucidation of relationships between objects and attributes through graph representations extracted from images.
In this paper, we extend the Ego4D dataset~\cite{grauman2022ego4d} with the proposed graph-based egocentric action annotations, with the goal of enhancing long-form video understanding and enabling further investigations on the use of graph-based representations in egocentric vision.

\section{Egocentric Action Scene Graphs}

Egocentric Action Scene Graphs (EASGs) provide annotations for a video clip in the form of a dynamic graph.
We formalize an EASG as a time-varying directed graph $G(t) = (V(t),E(t))$, where $V(t)$ is the set of nodes at time $t$ and $E(t)$ is the set of edges between such nodes (Figure~\ref{fig:notation}). Each temporal realization of the graph $G(t)$ corresponds to an egocentric action spanning over a set of three frames defined as in~\cite{grauman2022ego4d}: the \textit{precondition} (PRE), the \textit{point of no return} (PNR) and the \textit{postcondition} (POST) frames. The graph $G(t)$ is hence effectively associated to three frames: $\mathcal{F}(t) = \{PRE_t, PNR_t, POST_t\}$. $G(t)$ has two fixed nodes: the camera wearer node $v_{cw}(t)$ representing the camera wearer, and the verb node $v_{verb}(t)$, describing the action performed by the camera wearer at time $t$. Each graph $G(t)$ also contains a set of object nodes $V_{obj}(t)$ encoding the objects involved in the actions. In this formulation, the camera wearer's hands will appear as object nodes. In sum, we have: $V(t) = \{v_{cw}(t), v_{verb}(t)\} \cup V_{obj}(t)$. Apart for the camera wear node, each other node is associated to one or more attributes through a function $att$. 
Hence, for the camera wearer node, we define $att(v_{cw}(t)) = \varnothing$. The verb node is associated to a \textit{verb class attribute}: $att(v_{verb}(t)) = verb$. Noun nodes $v_{i}(t)$ are associated to a \textit{noun class attribute} $noun$ and to three bounding box attributes grounding the noun to the $PRE(t)$, $PNR(t)$ and $POST(t)$ frames associated to the action taking place at time $t$: $att(v_{i}(t)) = (noun, box_{PRE}, box_{PNR}, box_{POST})$. Note that two nodes indexed by the same subscript $i$ are related to the same physical object instance regardless of time~$t$. For instance $v_i(t)$ and $v_i(t')$ represent the same physical objects even when $t \neq t'$, but their associated bounding box attributes may not correspond as they are related to different frames.

The edges in the graph describe the relationships between nodes. Let $v_i(t)$ and $v_j(t)$ be two nodes in the graph. Then, we can define an edge $(v_i(t), v_j(t)) \in E(t)$ if there is a relationship between the nodes $v_i(t)$ and $v_j(t)$ at time $t$. We represent the existence of an edge between nodes $v_i(t)$ and $v_j(t)$  using the function $e_t$, such that $e_t(v_i(t),v_j(t))=r$, if there is a relationship $r$ between nodes $v_i(t)$ and $v_j(t)$; and $e_{t}(v_i(t),v_j(t)) = \varnothing$ otherwise. We require $r \in R$, where $R$ is the set of possible relationships between nodes.
Relations between verb and object nodes can be of a \textit{direct object} kind (e.g., puts -- \textit{dobj} --package), or a preposition (i.e., puts -- \textit{in} -- fridge), while relationships between object nodes are characterized by the prepositions only (i.e., package --\textit{with} -- carrot). Objects  $v_i(t)$ which are in a \textit{direct object} relation with the verb node $v_{verb}(t)$ are also referred to as ``direct objects'', while all other objects are referred to as ``indirect objects''. There is always an \textit{action} relationship between $v_{cw}(t)$ and $v_{verb}(t)$, i.e., $(v_{cw}, v_{verb}) \in E(t) \land r(v_{cw}, v_{verb})=action$.

Since our representation is centered on the action currently executed by the camera wearer, we add only the objects that are either direct objects (e.g., objects manipulated by $v_{cw}(t)$), or objects that have a direct relationship with either the verb node or any direct object nodes. For example, if the \textit{camera wearer} takes an \textit{apple} from the \textit{table} on which many other objects are located (e.g., a pear), only the apple and table will appear as nodes of the EASG, whereas \textit{pear} will not. 
\section{Ego4D-EASG Dataset}\label{sec:dataset}

We build our EASG dataset, \textit{Ego4D-EASG}, by annotating a subset of $221$ Ego4D~\cite{grauman2022ego4d} clips sampled over $181$ distinct videos
containing labels for the State Change Object Detection benchmark (SCOD). These labels, together with the narrations available in Ego4D are used to seed the collection of EASG annotations.
Let $\mathcal{C} = \{C_1, \ldots, C_N\}$ be the set of selected clips. Each clip $C_i$ consists in a sequence of object state change annotations from the SCOD benchmark $C_i=\{a_{t_1}^i, a_{t_2}^i, \ldots, a_{t_{m_i}}^i\}$, where the generic annotation $a_t^i = (a_t^{i,PRE}, a_t^{i,PNR}, a_t^{i,POST})$ contains annotations for three salient frames related to an object-state change at time $t$ of the clip $C_i$: the precondition (PRE), the point of no return (PNR), and the postcondition (POST). Each annotation is defined as $a_t^{i,x} = (f, n, b_o, b_{lh}, b_{rh}, r)$, where $x$ is either PRE, PNR or POST, $f$ is the frame number, $n$ and $b_o$ are the noun class and bounding box of the object of change (the manipulated object), $b_{lh}$ is the bounding box of the left hand, $b_{rh}$ is the bounding box of the right hand, and $r$ is a corresponding free-form narration which is matched to the current annotation. If the right or left hands are not visible in the scene, then either $b_{lh}=\varnothing$ or $b_{rh}=\varnothing$.
We labeled an independent EASG $G_i(t)$ for each clip $C_i$. Each temporal realization of the graph, $G_i(t)$ is seeded from the annotation tuple $a_t^i = (a_t^{i,PRE}, a_t^{i,PNR}, a_t^{i,POST})$. 
The data annotation is performed in two stages: 1) the graph annotation stage, and 2) the graph validation stage. These two stages are detailed in the following sections. We used Amazon Mechanical Turk for both stages. After data graphs annotation and validation, a temporal recollection stage allows to turn individual graphs into temporal dynamic graphs. The annotation process is discussed in the following sections. We will release the code to collect annotations following the proposed procedure (GitHub page: \href{https://github.com/fpv-iplab/EASG}{https://github.com/fpv-iplab/EASG}).

\begin{figure*}[tbp]
\centering
\includegraphics[width=\textwidth]{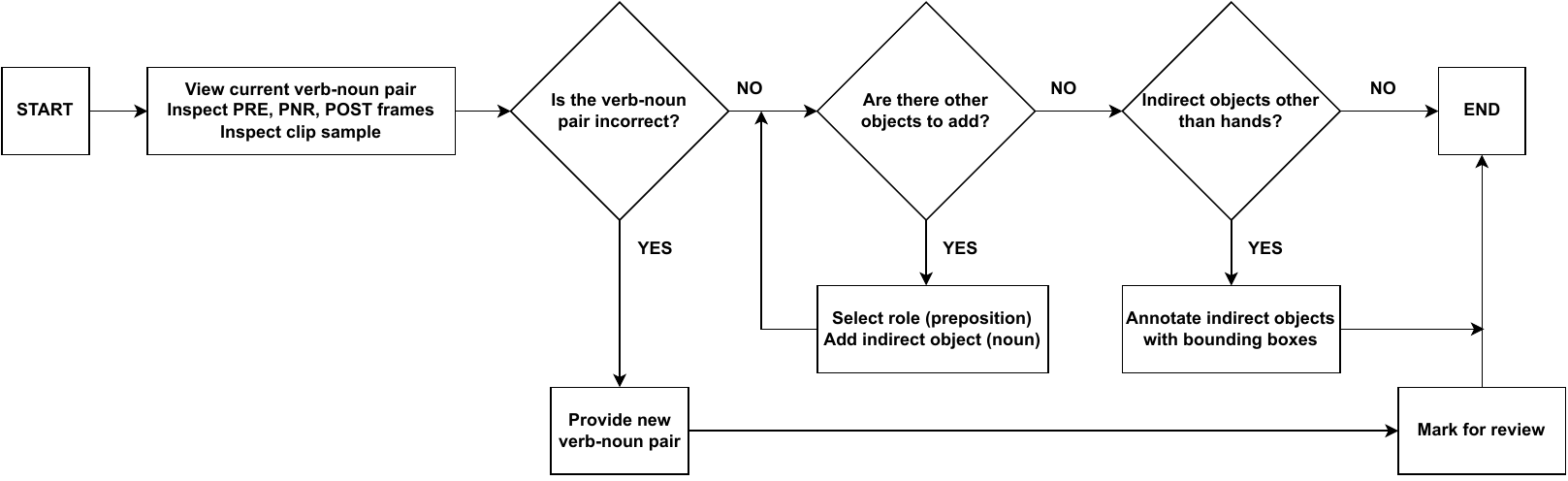} 
\caption{The Ego4D-EASG annotation pipeline. The annotators first review the provided verb-noun pair, the $PRE$, $PNR$, $POST$ frames and a clip sampled around $PNR$. They then check the existing narration and add indirect objects and related groundings, if necessary.}
\label{fig:pipeline}
\end{figure*}

\subsection{Egocentric Action Scene Graph Annotation}
This stage aims to obtain initial EASG $G_i(t)$ from annotations $a_t^i \in C_i$. This is done through an initialization and a refinement procedures.

\noindent
\textbf{Graph Initialization} We add by default the camera wearer node $v_{cw}(t)$, the verb node $v_{verb}(t)$, and set the default \textit{action} edge $e_t(v_{cw}(t),v_{verb}(t))=action$. The \textit{verb} attribute of $v_{verb}(t)$ is set by extracting the verb belonging to the narration $r$ associated to the current annotation $a_t^i$. We then initialize a new object node $n_k(t)$ to represent the manipulated object. We set the \textit{noun} and \textit{box} attributes of $n_k(t)$ as the noun $n$ and bounding box $b_o$ annotations included in $a_t^i$ ($n, b_o \in a_t^i$). We add a \textit{direct object} edge between $v_{verb}(t)$ and $n_k(t)$: $e_t(v_{verb}(t),n_k(t))=direct\ object$. 

\noindent
\textbf{Graph Refinement}
We ask three independent AMT annotators to provide manual annotations in order to refine the initial graph $G_i(t)$. The annotation pipeline for this stage is shown in the Figure~\ref{fig:pipeline}. We first ask annotators to inspect the provided verb-noun pair, the associated $PRE$, $PNR$ and $POST$ frames and a video clip of $5$ seconds sampled around the $PNR$ frame. Initial verb-noun pairs are obtained extracting the verb from the narration and the noun from the SCOD annotation. Annotators then check if the verb-noun pair corresponds to the observed clip; if it does not, then the annotators provide a correct $(verb,noun)$ pair and the current annotation is ended and marked for later review. We observe that narrations do not match in $<5\%$ of the cases. These examples have been later manually checked and re-labeled following the same procedure. We then ask the annotators to specify and ground any additional objects which may be linked to the verb node $v_{verb}$ or any existing object nodes. Note that only indirect objects can be added in this stage. For each newly added object node $v_{k}$, annotators are also asked to specify the preposition linking this object to the current graph (e.g., ``the camera wearer takes bowl \textbf{with} right hand'', where ``right hand'' is the new object and ``with'' is the specified preposition). We prompt the annotators with some likely objects which may appear in the frame and related prepositions, extracted from the narration $r$ through part of speech tagging, but the annotators were free to add any new objects they may find relevant in the observed video clip. For each of the added \textit{indirect} objects, annotators are also asked to ground them to the $PRE$, $PNR$, and $POST$ frames through bounding boxes. If the added objects correspond to the hands, the groundings are set to $b_{lh}$ and $b_{rh}$ as specified in the annotation $a_j$. At the end of this process, we obtain three graphs $G_i^1(t)$, $G_i^2(t)$, $G_i^3(t)$ as labeled by the three independent annotators.



\begin{table*}[htbp]
    \centering
    \footnotesize
    \begin{tabular}{p{5cm}p{6.5cm}c}
    \cline{1-2}
    \textbf{Validation procedure} & \textbf{Example Questions (\textcolor{red}{answers in red})} &  \textbf{Example Frame}\\
    \cline{1-2}
    1. Filtering verb-noun pair & Does CW \textit{take bowl} or \textit{press dough}? \textcolor{red}{take bowl} & \multirow{4}{*}{
    \includegraphics[width=4.5cm]{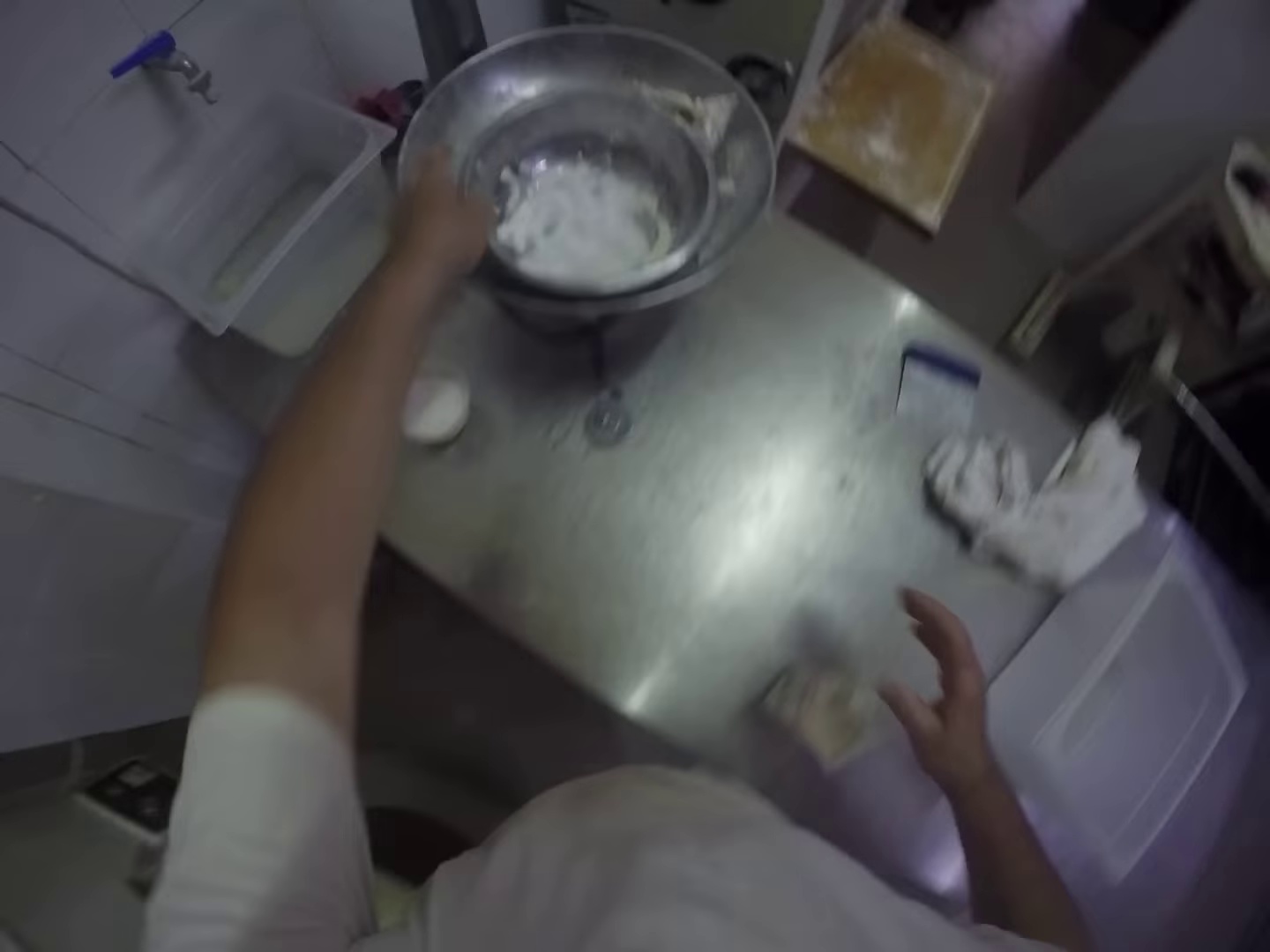}}\\ \cline{1-2}
    2. Selecting proper preposition in case of multiple edges between two nodes \vfill & \begin{minipage}{5.5cm} \vspace{3mm} Select the preposition which is more appropriate: \newline $\bullet$ CW takes bowl \textbf{with} left hand \textcolor{red}{\checkmark} \newline $\bullet$ CW takes bowl \textbf{on} left hand \end{minipage} \\ \cline{1-2}
    
    3. Selecting hand(s) if there are different hands with the same preposition & Does CW take bowl with right hand, with left hand or with both hands? \textcolor{red}{left hand}\\ \cline{1-2}
    
    4. Identifying spatial relations & \begin{minipage}{5.5cm} Is the following statement correct: \newline
    $\bullet$ The bowl is with flour [Y/N] \textcolor{red}{Y} \newline
    $\bullet$ The bowl is from scale [Y/N] \textcolor{red}{N} \end{minipage}\\
    \cline{1-2}
    \end{tabular}
    \caption{Examples of questions (with correct answers in red) asked to the annotators in the validation stage to resolve ambiguities between the labels provided in the annotation stage.}
    \label{tab:validation_stage}
\end{table*}

\subsection{Egocentric Action Scene Graph Validation}
The validation stage aggregates the data received from the three annotators and ensures the quality of the final annotations. In this stage, for each ($G_i^1(t)$, $G_i^2(t)$, $G_i^3(t)$) graph tuple, we show the annotators the $PRE$, $PNR$, and $POST$ frames, the video clip sampled around the $PNR$ and ask a set of questions aiming to sort out inconsistencies across the three graphs. We formulate up to four questions: 1) a question aimed to select the correct verb-noun pair if there is disagreement in the three graphs; 2) a question aimed to disambiguate relations between pairs of nodes, if the three graphs have disagreeing edges between the same node pairs; 3) a question aimed to identify the correct hand used to manipulate objects; 4) a question aimed to disambiguate spatial relationships. The answers provided by the annotators to each of these questions allow to resolve ambiguities and obtain a single graph $G_i(t)$ for each clip $C_i$ and each timestamp $t$. Table~\ref{tab:validation_stage} reports example questions and correct answers for an example annotation.

\subsection{Temporal Recollection}
The graphs $G_i(t)$ obtained through the annotation and validation stages are \textit{static graphs}, meaning that node indices at different timestamps do not necessarily indicate the same object. For instance, the object ``plate'' may be identified by $n_i(t)$ and $n_j(t')$ with $t \neq t'$. In this stage, we reason globally on the dynamic graph $G_i(t), t=1\ldots,T$ and re-assign node indices to make sure that object nodes representing the same object instance are assigned the same index. At the end of this process, a ``plate'' object will be indexed with the same subscript across timestamps: $n_i(t)$ and $n_i(t')$ with $t \neq t'$. This makes sure that $G_i(t)$ can be interpreted as a dynamic graph across all timestamps.

\begin{figure*}[tbp]
\centering
\includegraphics[width=0.27\textwidth, height=0.2\textwidth]{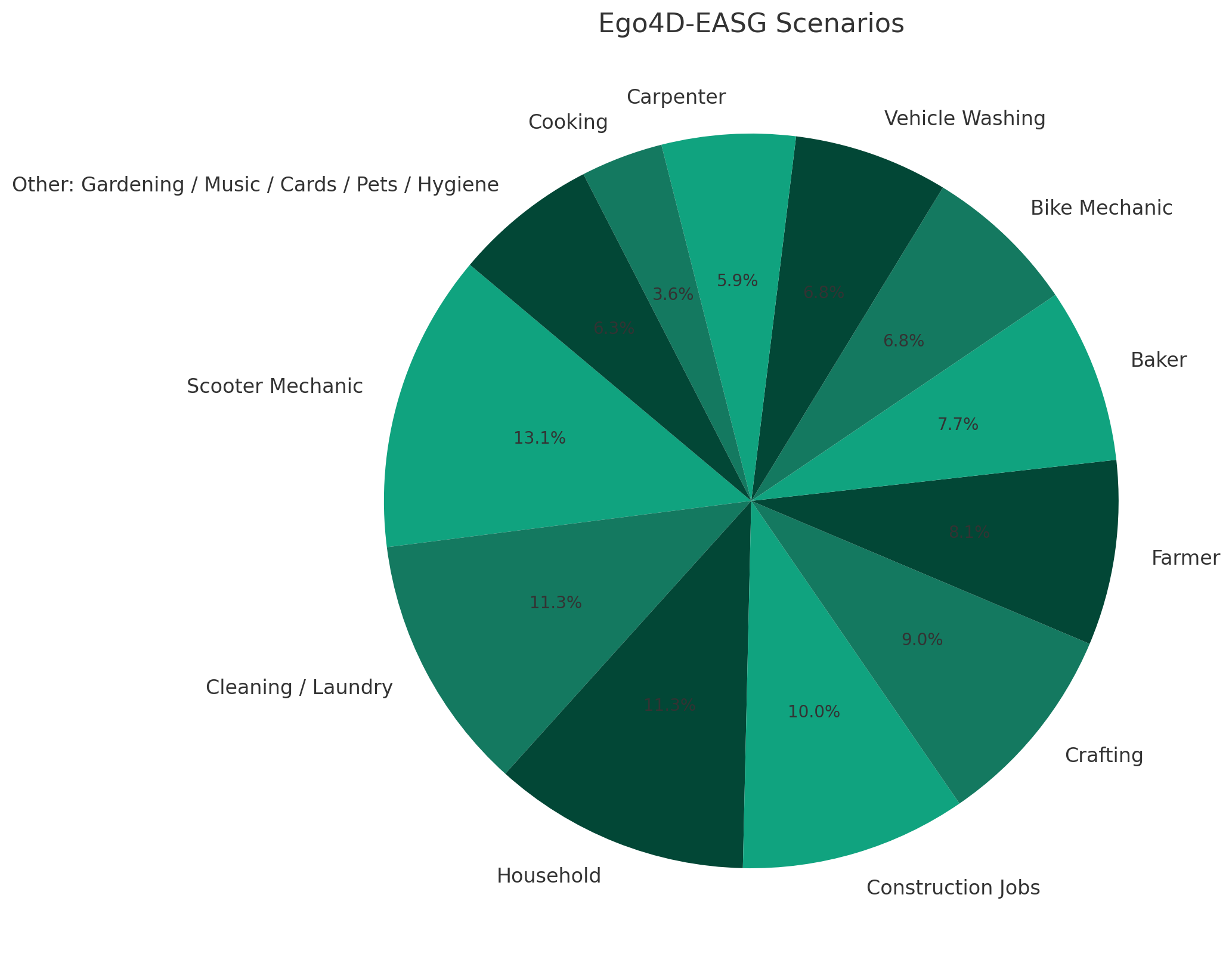}
\hfill
\includegraphics[width=0.6\textwidth, height=0.2\textwidth]{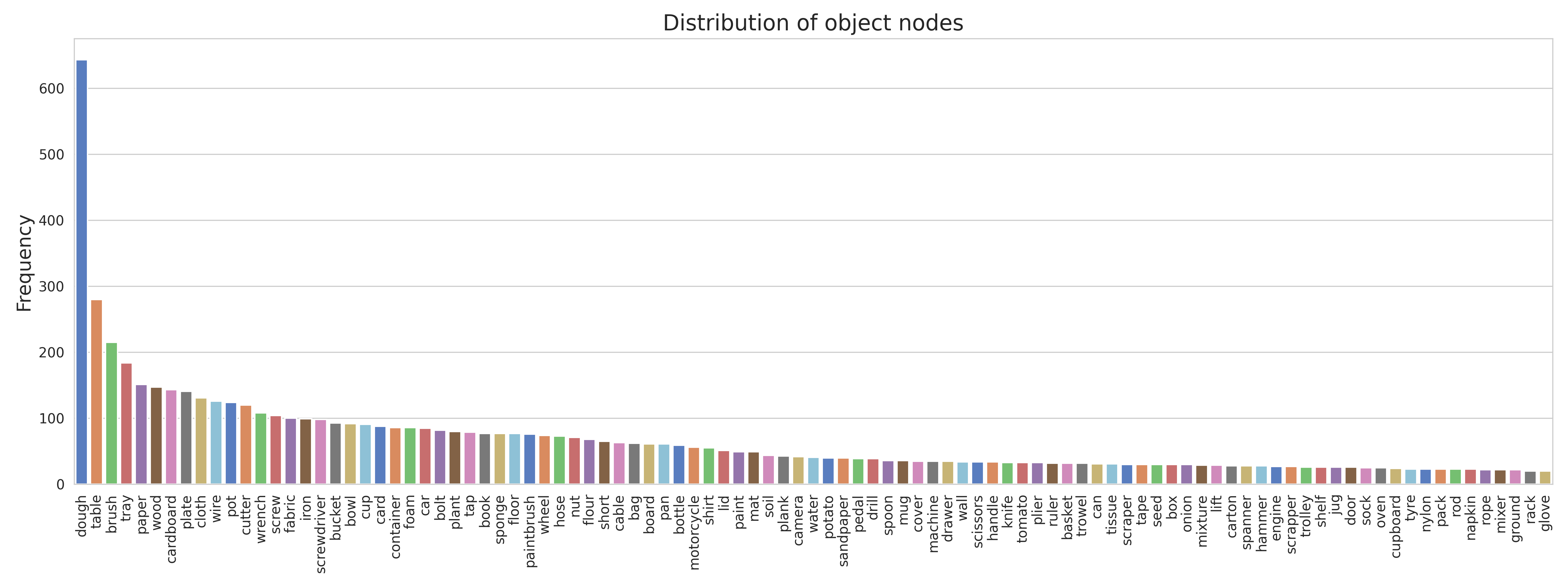}

\vspace{2mm}
\includegraphics[width=0.29\textwidth, height=0.18\textwidth]{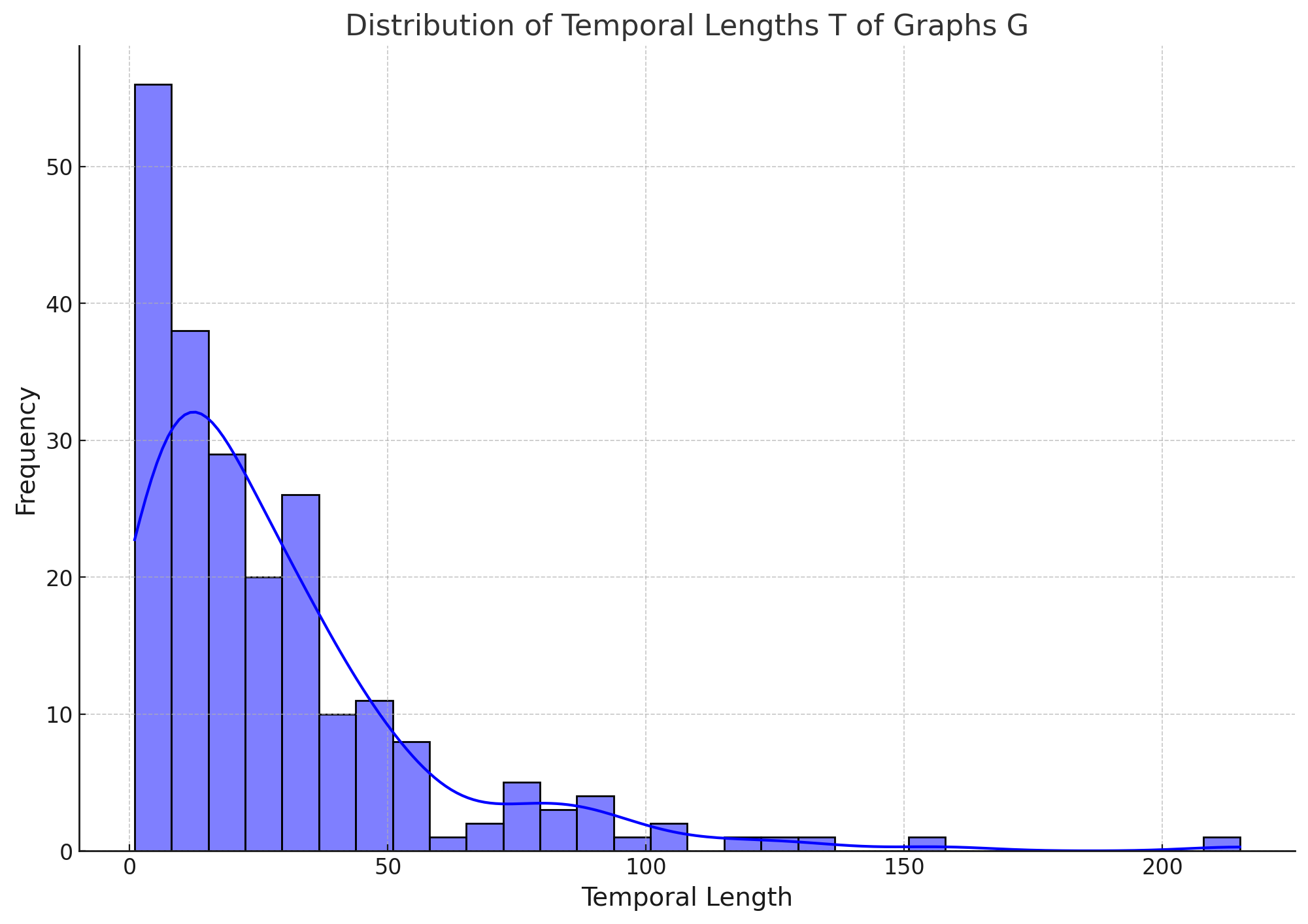}
\hfill
\includegraphics[width=0.44\textwidth, height=0.18\textwidth]{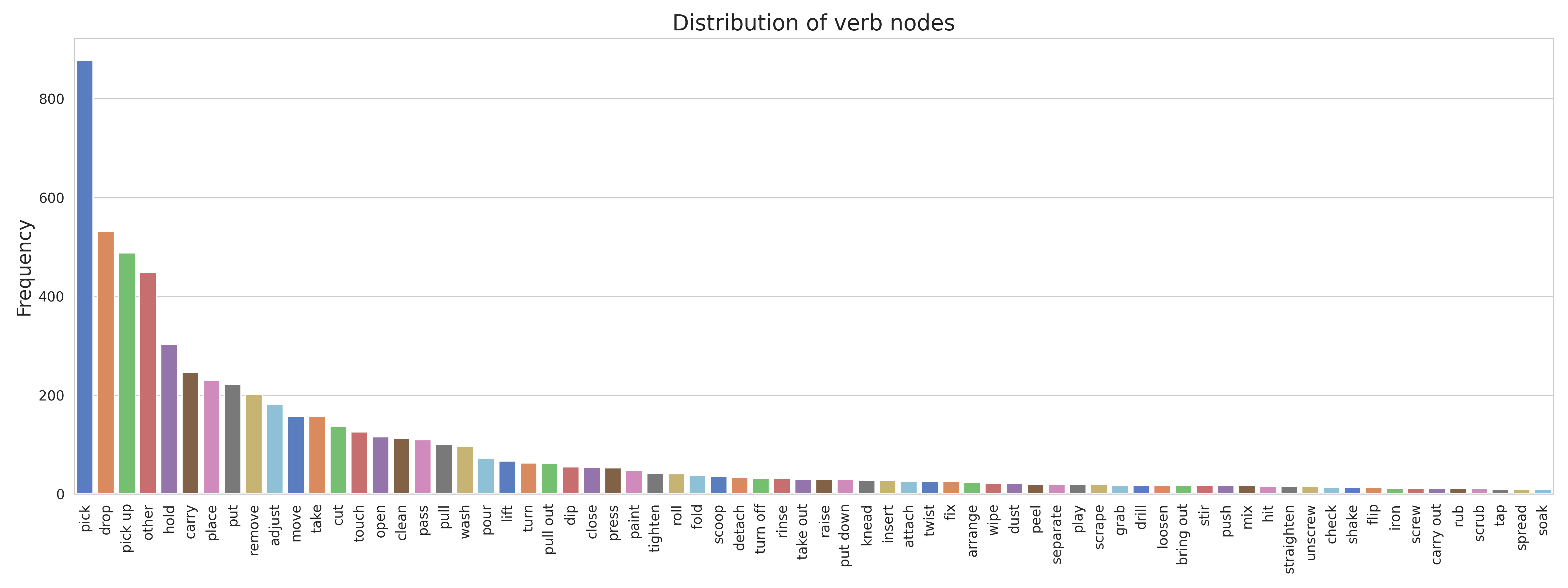}
\hfill
\includegraphics[width=0.25\textwidth, height=0.18\textwidth]{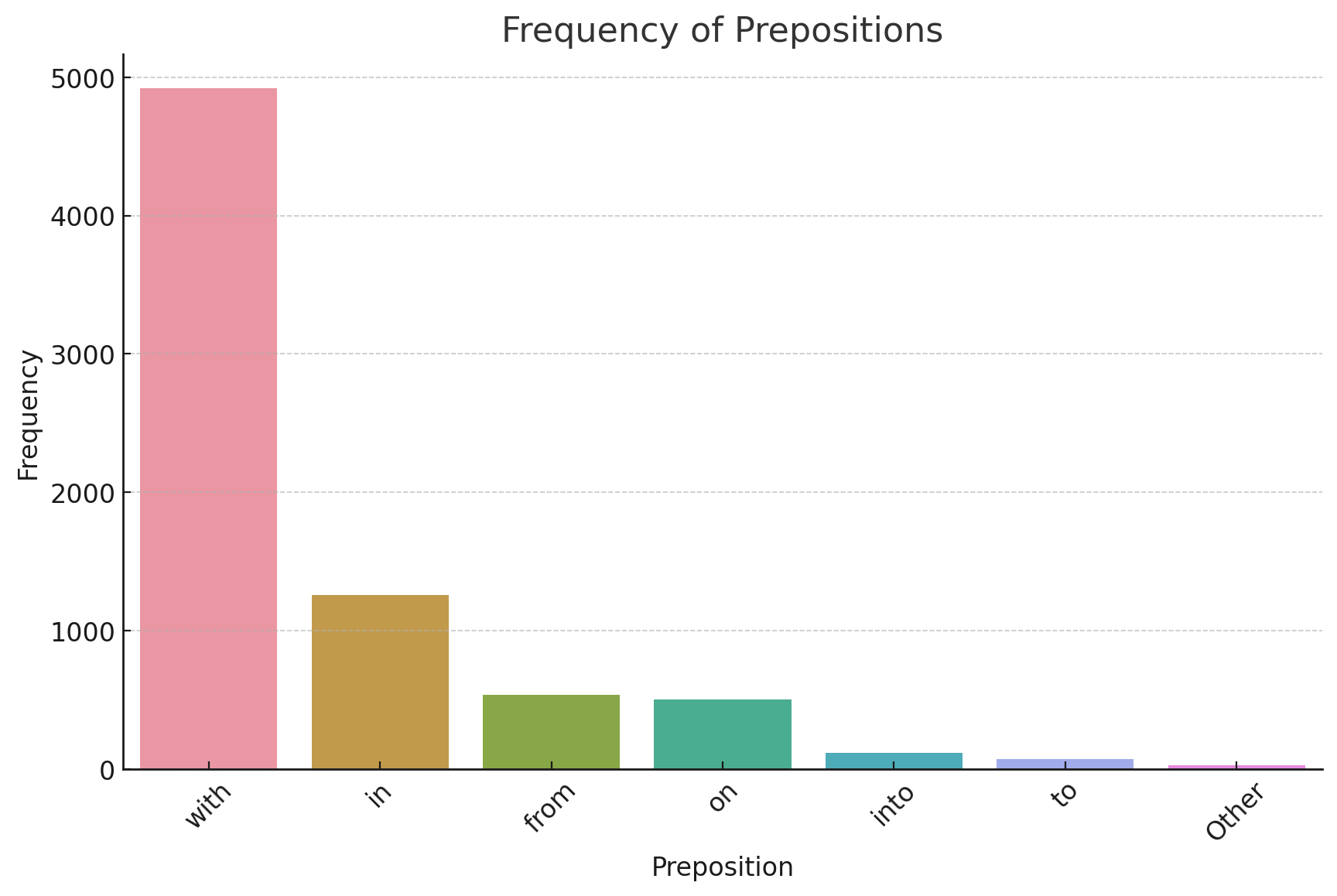}
\caption{Left-to-right, top-to-bottom: Distributions of clips across scenarios, object nodes, temporal lengths $T$ of graphs $G$, verb nodes, and relation categories (excluding \textit{action} and \textit{direct object} relations). Data is distributed across different scenarios related to egocentric perception, long-tailed object, verb distributions, and prepositions. The distribution of temporal length of graphs shows the long-form nature of our annotations, with most graphs having a length of up to $50$ timesteps.
}
\label{fig:statistics}
\end{figure*}


\subsection{Dataset Statistics and Comparison with Other Scene Graph Datasets}
Table~\ref{tab:statistics_comparison} reports statistics on the proposed Ego4D-EASG dataset and compares it with existing video scene graph datasets. The proposed dataset is the only one designed for long-from egocentric video understanding and it features $221$ egocentric video sequences, 11.4 hours of video\footnote{We measure the length of each sequence from the timestamp of the $G(1):PRE$ frame to the timestamp of the $G(T):POST$ frame.}, comprising an average labeled sequence length of 3.1 minutes, $T = 28.3$ graphs per video in average, $407$ object classes, $219$ verb classes, and $16$ relation classes. As compared to previous datasets, ours is the only including verb nodes explicitly encoding actions. As a result, the number of relations, which in previous datasets also encoded actions (e.g., ``looking at'') is lower than in other datasets. 

\begin{table*}[]
\resizebox{\linewidth}{!}{%
    \centering
    \begin{tabular}{l|cc|ccccccc}
    \hline
         \textbf{Dataset} & \textbf{Dynamic} & \textbf{Egocentric} & \textbf{Sequences} & \textbf{Hours} & \textbf{Avg. Len. (seconds)} & \textbf{Avg. Graphs per Vid.} &  \textbf{Obj Cls} & \textbf{Verb Cls} & \textbf{Rel Cls}\\
         \hline
         VidVRD~\cite{VidVRD} & \xmark & \xmark & 1,000 & 3 & 11 & 3.9* & 35 & 25** & 132\\
    VidOR~\cite{VidOR} & \xmark & \xmark & 10,000 & 99 & 35 & 8.8* action + 29.2* spatial & 80 & 42 & 50\\
    Action Genome~\cite{wu2020adaptive} & \checkmark & \xmark & 10,000 & 82 & 30 & 5 & 35 & - & 25\\
    PVSG~\cite{PVSG2023} & \xmark & Partly (28\%) & 400 & 9 & 77 & 382 & 126 & 44 & 57\\
    HOMAGE~\cite{rai2021home} & \xmark & paired ego-exo & 1,752 & 25 & 3 & 3.8 & 86 & 453 & 29\\
    \hline
    Ego4D-EASG (Ours) & \checkmark & \checkmark & 221 & 11.4 & 186 & 28.3 & 407 & 219 & 16 \\
    \hline
    \end{tabular}
    }
    \caption{Comparison with existing video scene graph datasets. Our Ego4D-EASG dataset is the only one explicitly designed for long-from egocentric video understanding, featuring egocentric videos, dynamic graphs, an average sequence length of 3.1 minutes and an average number of 28.3 graphs per sequence. *measured in object-relation-object triplets. **intransitive + transitive verb predicates.}
    \label{tab:statistics_comparison}
\end{table*}

Out of the all clips, $129$ belong to the SCOD-train split and $92$ to SCOD-val slit. The dataset contains 30,478 and 19,342 bounding boxes as object groundings in train and validation splits respectively. For an exhaustive enumeration of the sets of all verbs $V_{verb}$, objects $V_{obj}$, and relations $R$, please refer to the supplementary material. 
Figure~\ref{fig:statistics} reports statistics on the distribution of scenarios, nouns, verbs, relations, and temporal graph lengths.
\section{Egocentric Action Scene Graphs Generation}\label{sec:generation}

\begin{table*}[tb!]
\centering
\vspace{-0.5em}
\setlength{\tabcolsep}{3pt}
\resizebox{1\linewidth}{!}{%
\begin{tabular}{lcccccccccccccccccc}
\toprule
  \multirow{3}[3]{*}{Method} & \multicolumn{9}{c}{With Constraint} & \multicolumn{9}{c}{No Constraint}\\
  \cmidrule(lr){2-10} \cmidrule(lr){11-19} 
   & 
 \multicolumn{3}{c}{\textit{Edge Cls}} & \multicolumn{3}{c}{\textit{SG Cls}} & \multicolumn{3}{c}{\textit{EASG Cls}} & \multicolumn{3}{c}{\textit{Edge Cls}} & \multicolumn{3}{c}{\textit{SG Cls}} & \multicolumn{3}{c}{\textit{EASG Cls}} \\ 
    \cmidrule(l){2-4} \cmidrule(l){5-7}  \cmidrule(l){8-10} \cmidrule(l){11-13} \cmidrule(l){14-16} \cmidrule(l){17-19}  
  &R@10 &R@20 &R@50 &R@10 &R@20 &R@50 &R@10 &R@20 &R@50 &R@10 &R@20 &R@50 &R@10 &R@20 &R@50 &R@10 &R@20 &R@50
\\
\midrule
\midrule
\\[-1.1em]
  Random Guess & 8.0 & 8.0 & 8.0 & 0.2 & 0.4 & 1.0 & 0.0 & 0.0 & 0.0 & 36.5 & 72.6 & 99.9 & 0.3 & 0.5 & 1.0 & 0.0 & 0.0 & 0.0 \\
  Baseline (Ours) & 60.4 & 60.4 & 60.4 & 41.4 & 44.3 & 50.6 & 14.3 & 16.4 & 17.9 & 94.4 & 99.8 & 100 & 51.6 & 58.2 & 62.4 & 14.7 & 18.3 & 20.9 \\
\bottomrule
\end{tabular}
}
\caption{Baseline results for three EASG generation tasks (i.e. \textit{Edge Cls}, \textit{SG Cls}, and \textit{EASG Cls}) in terms of Recall@K.}
\label{tab:R_EASG_generation}
\end{table*}


\noindent\textbf{Task Definition} Unlike standard scene graph generation tasks, EASG generation aims to predict the action verbs as well as objects and their relationships. To this end, we define three EASG generation tasks as follows: (1) Edge classification (\textit{\mbox{Edge Cls}}) is to predict verb-object and object-object relationships given visual features, the ground-truth action verb and object classes, (2) Scene Graph Classification (\textit{\mbox{SG Cls}}) is to predict both the object classes and the edge relationships given visual features and the ground-truth action verb, and (3) Egocentric Action Scene Graph Classification (\textit{\mbox{EASG Cls}}) is to predict all these three components, which encompass action verbs, objects, and edge relationships. 

\noindent\textbf{Experimental Setting} As EASG is a new type of egocentric video representation, it is not trivial to apply previous approaches of image or video scene graph generation to the EASG tasks. Therefore, we design an original baseline model consisting of task-specific fully-connected layers working on top of pre-extracted visual features. For \textit{\mbox{Edge Cls}}, we use a single-layer model to predict the edge relation from the clip-level features and ROIAlign features of each object bounding box. For the clip-level features, we take the average of SlowFast~\cite{feichtenhofer2019slowfast} features (pre-extracted and provided within the Ego4D dataset~\cite{grauman2022ego4d}) for the whole clip spanning from \textit{PRE} to \textit{POST} frames. We extract the ROIAlign features using the Faster-RCNN~\cite{ren2015faster} pre-trained for the short-term action anticipation benchmark~\cite{grauman2022ego4d}. For \textit{\mbox{SG Cls}}, we add an additional fully-connected layer to predict the object classes from the ROIAlign features. For \textit{\mbox{EASG Cls}}, we add another additional layer to predict the action verb from the clip-level features. Following the convention in the literature of scene graph generation, we evaluate this baseline under two different setups: \textit{With Constraint} and \textit{No Constraint}. The former restricts each graph to have at most a single verb-object relationship, whereas the latter has no such restriction. The baseline model is trained for 10 epochs using the Adam optimizer~\cite{kingma2014adam} with a learning rate of $10^{-3}$.

\noindent\textbf{Results} We report the baseline results for all different tasks and setups in Table~\ref{tab:R_EASG_generation} using the standard metrics of Recall@K (R@K, K=[10, 20, 50]). Baseline results are compared with random guess. We can observe that the scores of \textit{\mbox{EASG Cls}} are significantly lower than other results, indicating that action verbs introduce another layer of difficulty to EASG understanding.
\section{Downstream long-from video understanding tasks with Egocentric Action Scene Graphs}\label{sec:downstream}

In this section, we report experiments aimed to show the potential of the EASG representation in the downstream tasks of action anticipation and activity summarization. Both tasks require to perform long-form reasoning of egocentric video, processing long video sequences spanning over different timesteps.
Following recent results showing the flexibility of Large Language Models (LLMs) as symbolic reasoning machines~\cite{mirchandani2023large}, we perform these experiments with LLMs accessed via the OpenAI API \cite{openai2023gpt}. 
The experiments aim to examine the expressive power of the EASG representation and its usefulness for downstream applications. 
We show that EASG offers an expressive way of modeling long-form activities, in comparison with the gold-standard verb-noun action encoding, extensively adopted in previous work~\cite{damen2020rescaling, grauman2022ego4d}. 
The exact prompts used in the experiments and additional results are provided in the supplement. 

\subsection{Action anticipation with EASGs}

\textbf{Experimental Setting} For the action anticipation task, we use the GPT3~\cite{brown2020language} \textit{text-davinci-003} model. 
We prompt the model to predict the future action from a sequence of length $T \in \{5,20\}$.
We compare two types of representations - EASG and sequences of verb-noun pairs. 
The input sequence of graphs can be represented as $s_{EASG} = [G(t_0), G(t_0+1),...,G(t_0+T-1)]$, with $t_0+T-1\geq 20$. 
Each graph $G(t)$ is represented as a string of triplets, where each triplet encapsulates the relationship between nodes (e.g., \textit{CW - verb - wash; wash - direct object - car; wash - with - sponge}).
As an output, we request to provide the future unobserved scene graph $G(t+T)$ in the same triplet format. From the predicted graph, we extract the action as the pair of verb and direct object node class for evaluation.
In the verb-noun baseline, the input sequence is represented as $s_{vn} = [s_{vn}(t_0), s_{vn}(t_0+1),...,s_{vn}(t_0+T-1)]$, with $t_0+T-1\geq 20$.
The generic term of the sequence is a $(verb, noun)$ pair extracted from the EASG annotation, where $noun$ is the noun class of the direct object.
The ground truth future action is $s_{vn}(t_0+T)$. 
Given the uncertainty in forecasting future events, we prompt the LLM to output up to $N=5$ predictions, a standard practice in anticipation~\cite{damen2018scaling, grauman2022ego4d}. We evaluate results using top-k accuracy, with $k \in \{1,5\}$, reported for verb, noun, and actions. The sample size for this experiment is $3030$.\\

\begin{table}[tbp]
\centering
\resizebox{\columnwidth}{!}{
\begin{tabular}{@{}lcccccccc@{}}
\toprule
& & & \multicolumn{2}{c}{Verb} & \multicolumn{2}{c}{Noun} & \multicolumn{2}{c}{Action} \\ 
\cmidrule(lr){4-5} \cmidrule(lr){6-7} \cmidrule(lr){8-9}
& Seq. length $T$ & Avg. duration & Top-1 & Top-5 & Top-1 & Top-5 & Top-1 & Top-5 \\
\midrule
V-N & 5 & 19s  & 2.54 & 5.01 & \underline{47.68} & 62.24 & 1.28 & 2.60 \\
EASG & 5 & 19s & 3.33 & \underline{9.53} & \textbf{48.84} & \underline{66.03} & 1.88 & \underline{5.24} \\
V-N & 20 & 82s & \underline{3.43} & 8.41 & 46.69 & 64.85 & \underline{2.01} & 4.98 \\
EASG & 20 & 82s & \textbf{5.94} & \textbf{15.97} & 47.36 & \textbf{67.26} & \textbf{3.40} & \textbf{9.24} \\
\hline
\multicolumn{3}{l}{Improvement}  & +2.51 & +7.56 & +0.67 & +2.41 & +1.39 & +4.26 \\
\bottomrule
\end{tabular}}
\caption{Performance Comparison for the Action anticipation task.}
\label{tab:anticipation}
\end{table}

\begin{figure*}
    \centering
    \includegraphics[width=\linewidth]{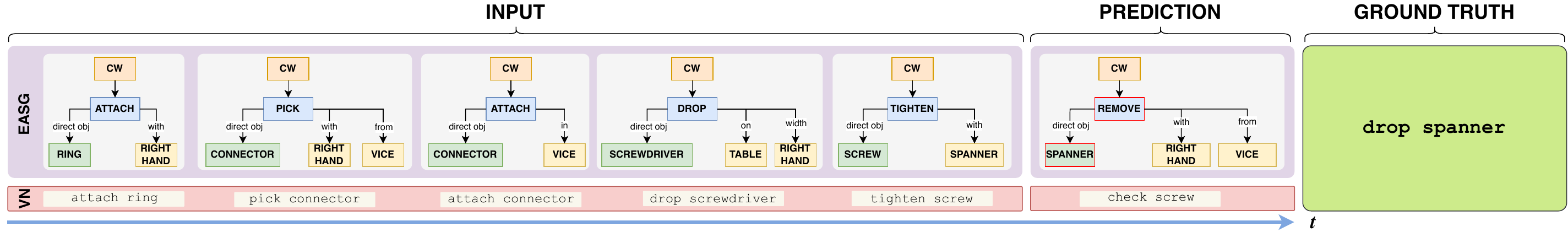}
    \caption{Qualitative example of input sequences and outputs produced using the EASG (top) and verb-noun (bottom) representations for action anticipation, along with the ground truth future action (right). The EASG prediction ``remove spanner'' is much more semantically aligned to the ground truth ``drop spanner'' action than ``check screw'', the prediction based on the verb-noun representation.}
    \label{fig:qualitative_anticipation}
\end{figure*}

\noindent\textbf{Results} Table~\ref{tab:anticipation} reports the results of these experiments. 
Best results are always achieved by EASG-based representations. As can be noted, even short EASG sequences ($T=5$) tend to outperform long V-N sequences ($T=20$), highlighting the higher representation power of EASG, when compared to standard verb-noun representations. EASG representations achieve the best results for long sequences ($T=20$). For instance, Top-5 verb is equal to $15.97$ for $T=20$, as compared to $9.53$ for $T=5$. These results further confirm the suitability of EASG for long-form understanding of egocentric video. EASGs bring overall significant improvements of up to $+7.56$ with respect to the best verb-noun based prediction across the different metrics. Figure~\ref{fig:qualitative_anticipation} reports a qualitative example.

\subsection{Long-form activity summarization with EASGs}

\textbf{Experimental Setting} We select a subset of $147$ Ego4D-EASG clips containing human-annotated summaries describing the activities performed in the clip in 1-2 sentences from Ego4D~\cite{grauman2022ego4d}.
We construct three types of input sequences: sequences of graphs $s_{EASG} = [G(1), G(2),...,G(T_{max})]$, sequences of verb-noun pairs $s_{vn} = [s_{vn}(1), s_{vn}(2),...,s_{vn}(T_{max})]$, and sequences of original Ego4D narrations, matched with the EASG sequence.
This last input is reported for reference, as we expect summarization from narrations to bring the best performance, given the natural bias of language models towards this kind of representation.
In this experiment, each $G(t)$ is represented as a sentence (e.g., \texttt{CW wash car with sponge}) to decrease the number of tokens in the input ($T_{max}$ can reach size up to 117) and to align with the natural language form of the predicted output.
We select clips for which $T_{max}\geq 5$. 
We use the \textit{GPT-3.5 Turbo} LLM model for these experiments.
We evaluate the produced summaries using the CIDEr~\cite{vedantam2015cider} metric, adopted in the image captioning literature, and standard metrics for NLG (ROUGE~\cite{lin2004rouge}, BLEU~\cite{papineni2002bleu}, METEOR~\cite{banerjee2005meteor}). \\

\begin{table}[t]
\centering
\resizebox{\linewidth}{!}{
\begin{tabular}{@{}lccccccccc@{}}
\toprule
 & CIDEr & ROUGE-1 & ROUGE-2 & ROUGE-L & BLEU-1 & BLEU-2 & BLEU-3 & BLEU-4 &  METEOR\\ 
\midrule
V-N & 9.42 & 31.5 & 10.3 & 29.7 & 35.7 & 18.6 & 7.6 & 3.9 &  26.09\\
EASG & 13.79 & 33.3 & 10.7 & 31.4 & 37.3 &  19.0 & 7.8 & 4.2 &   26.30\\
\hline
Narrations &19.99 & 37.7 & 14.0 & 34.4 &42.0 & 24.0 & 11.7 & 6.7 &   29.43\\
\bottomrule
\end{tabular}}
\caption{Results of activity summarization with EASGs and verb-noun representations.
}
\label{tab:summarization}
\end{table}

\noindent\textbf{Results} The results reported in Table~\ref{tab:summarization} indicate strong improvement in CIDEr score over $s_{vn}$ inputs, showing that models which process EASG inputs capturing detailed object-action relationships, will generate more specific, informative sentences that align well with reference descriptions. As expected, inputs based on narrations achieve the best performance. It should be noted that, while EASG are not as expressive as narrations, they provide a much more structured representation which may be beneficial for the development of computer vision systems. 
All the NLG metrics show improvements of $s_{EASG}$ over $s_{vn}$ representation, which indicates that indeed, Egocentric Action Scene Graphs provide meaningful information that can improve the quality of long-form video summarization.


\section{Conclusion}\label{sec:conclusion}
Our paper reports four key contributions. We introduced the Egocentric Action Scene Graphs (EASG) as a novel representation for understanding long-form egocentric videos. We established a procedure for the collection of such graphs and extended the Ego4D dataset with manually annotated EASG labels. We reported initial baseline results for EASG generation. We validated the EASG representation's effectiveness in two different downstream tasks, aimed at long-form egocentric video understanding. 
We believe that these contributions mark a step forward in long-form egocentric video understanding. To facilitate further research in this area, we will release the dataset and code, enabling replication of our data annotation and experimental methodologies (GitHub page: \href{https://github.com/fpv-iplab/EASG}{https://github.com/fpv-iplab/EASG}).

\paragraph{Acknowledgements.} This research is supported by Intel Corporation. Research at the University of Catania is supported in part by the project Future Artificial Intelligence Research (FAIR) – PNRR MUR Cod. PE0000013 - CUP: E63C22001940006.
{
    \small
    \bibliographystyle{ieeenat_fullname}
    \bibliography{main}
}

\clearpage
\appendix
\setcounter{page}{1}
\setcounter{section}{0}
\maketitlesupplementary

\section{Screenshots from the Annotation tool}
\label{sec:screenshots}
\begin{figure*}[!ht]
\centering
\begin{subfigure}[b]{0.49\textwidth}
    \centering
    \includegraphics[width=\linewidth]{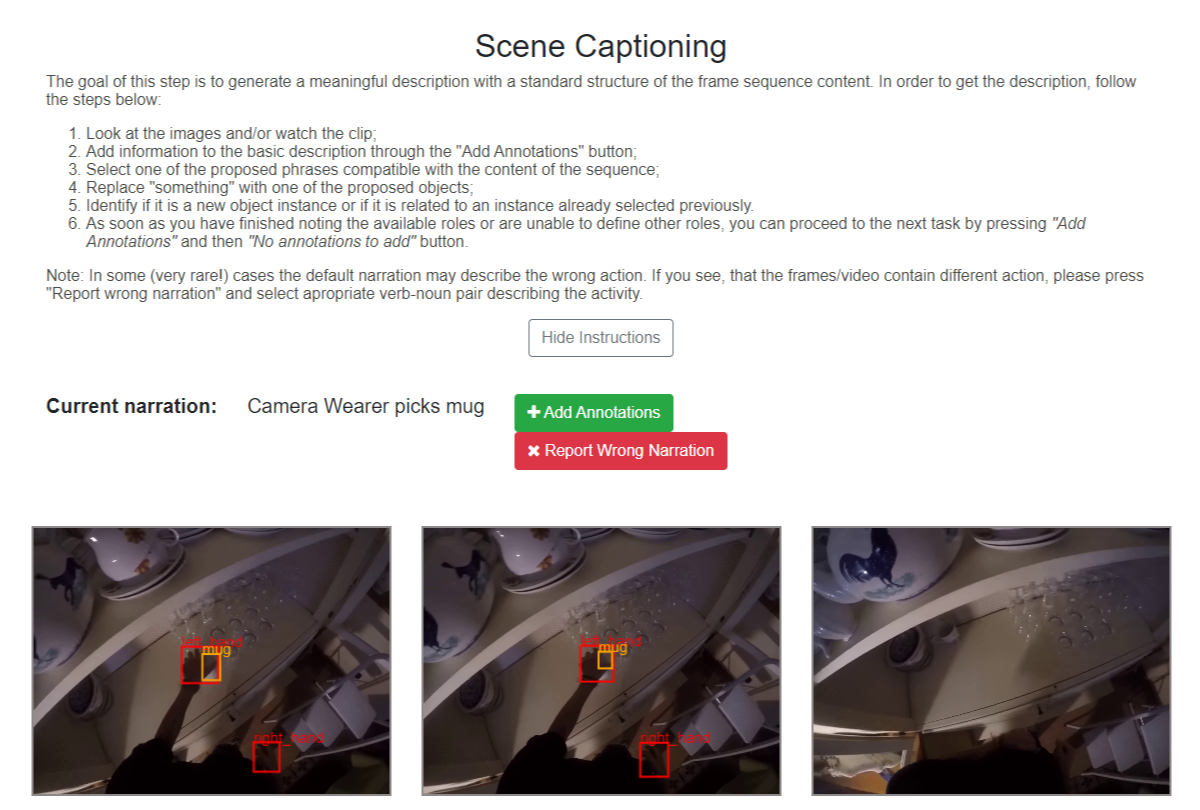}
    \caption{Initial interface}
    \label{fig:figure1}
\end{subfigure}%
\begin{subfigure}[b]{0.49\textwidth}
    \centering
    \includegraphics[width=\linewidth]{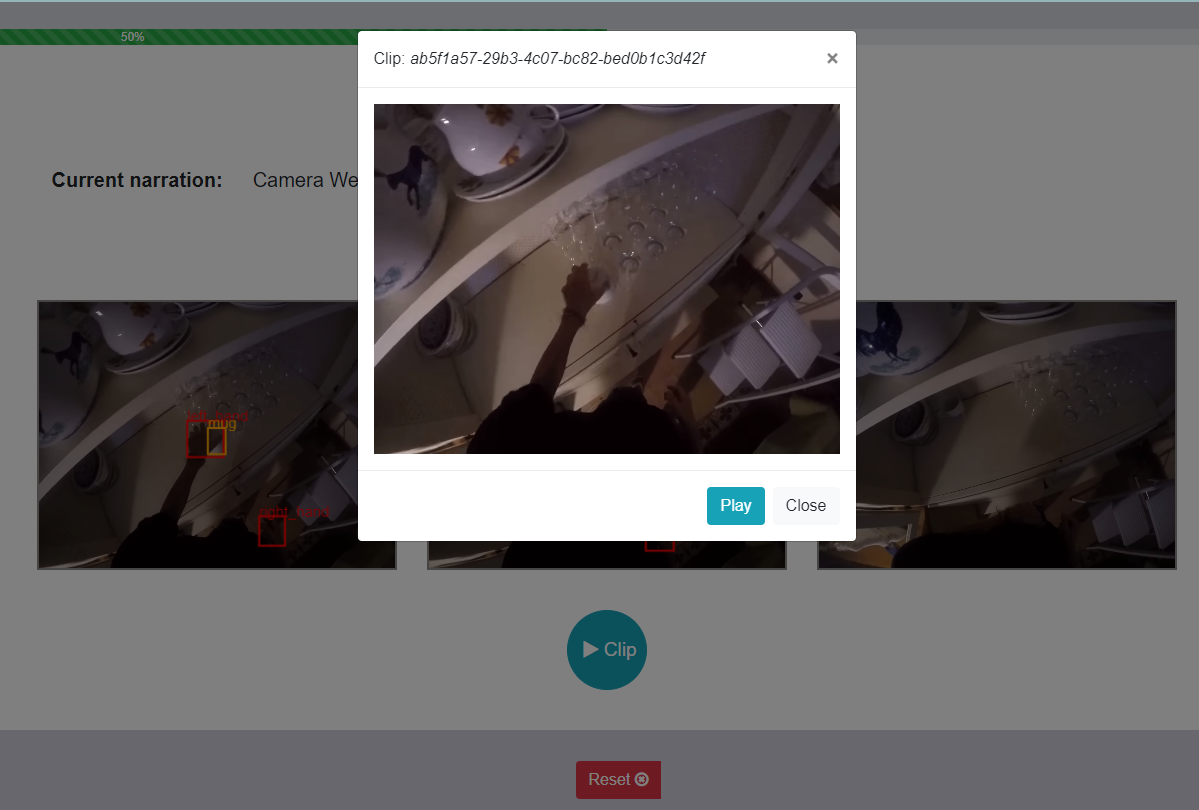}
    \caption{Clip sampled around PNR frame}
    \label{fig:figure2}
\end{subfigure}%

\begin{subfigure}[b]{0.49\textwidth}
    \centering
    \includegraphics[width=\linewidth]{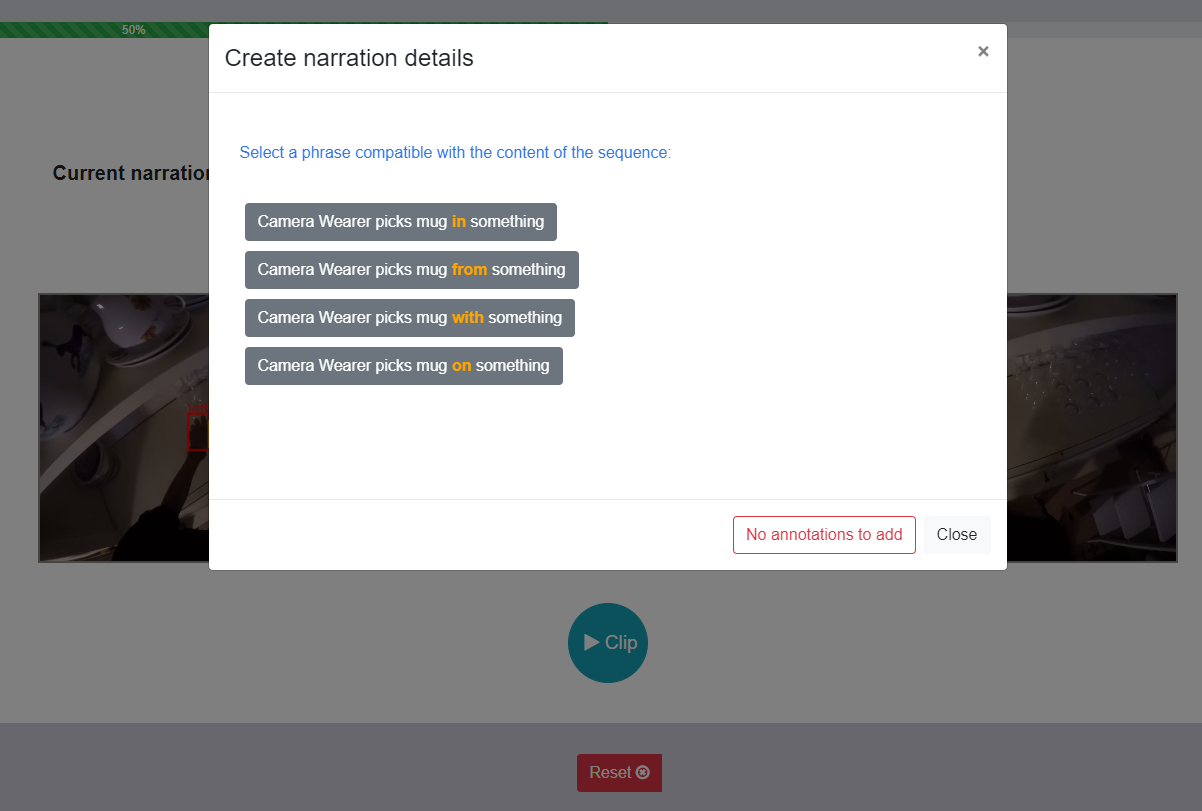}
    \caption{Selecting relation}
    \label{fig:figure3}
\end{subfigure}
\begin{subfigure}[b]{0.49\textwidth}
    \centering
    \includegraphics[width=\linewidth]{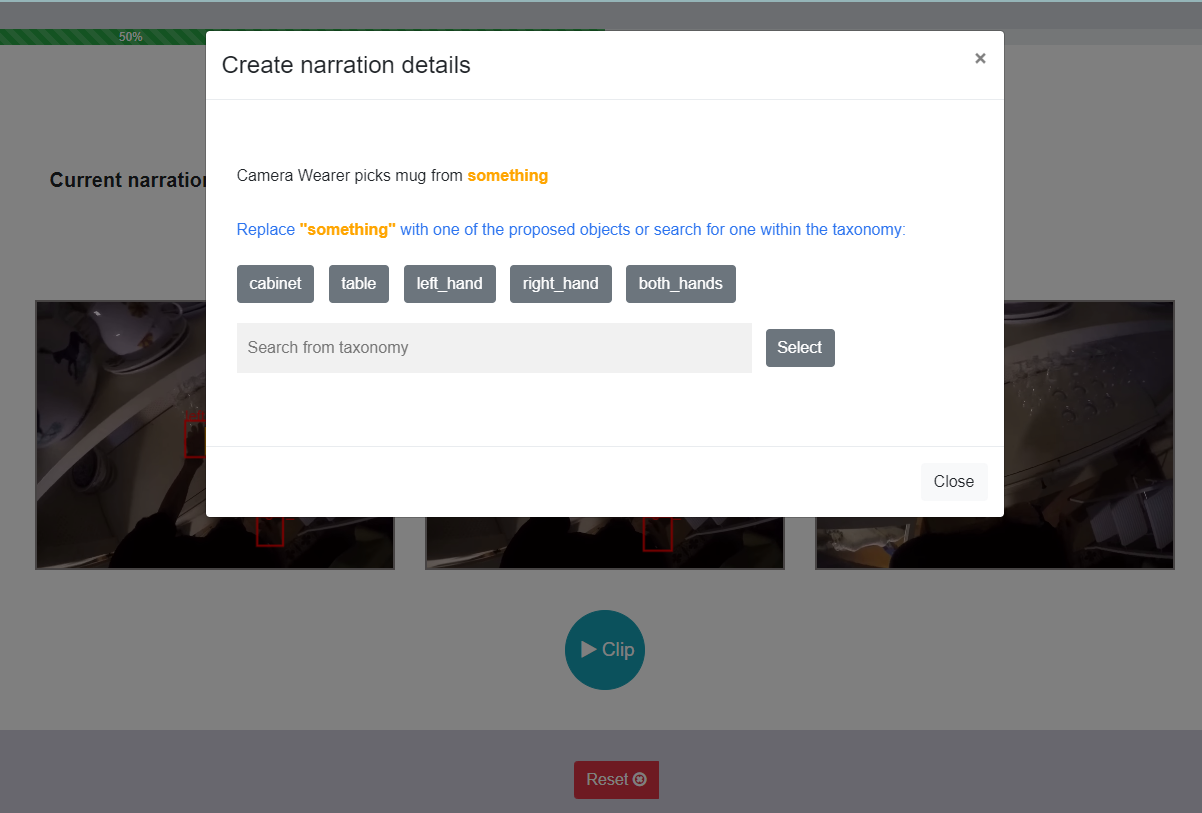}
    \caption{Adding indirect object}
    \label{fig:figure4}
\end{subfigure}%

\begin{subfigure}[b]{0.49\textwidth}
    \centering
    \includegraphics[width=\linewidth]{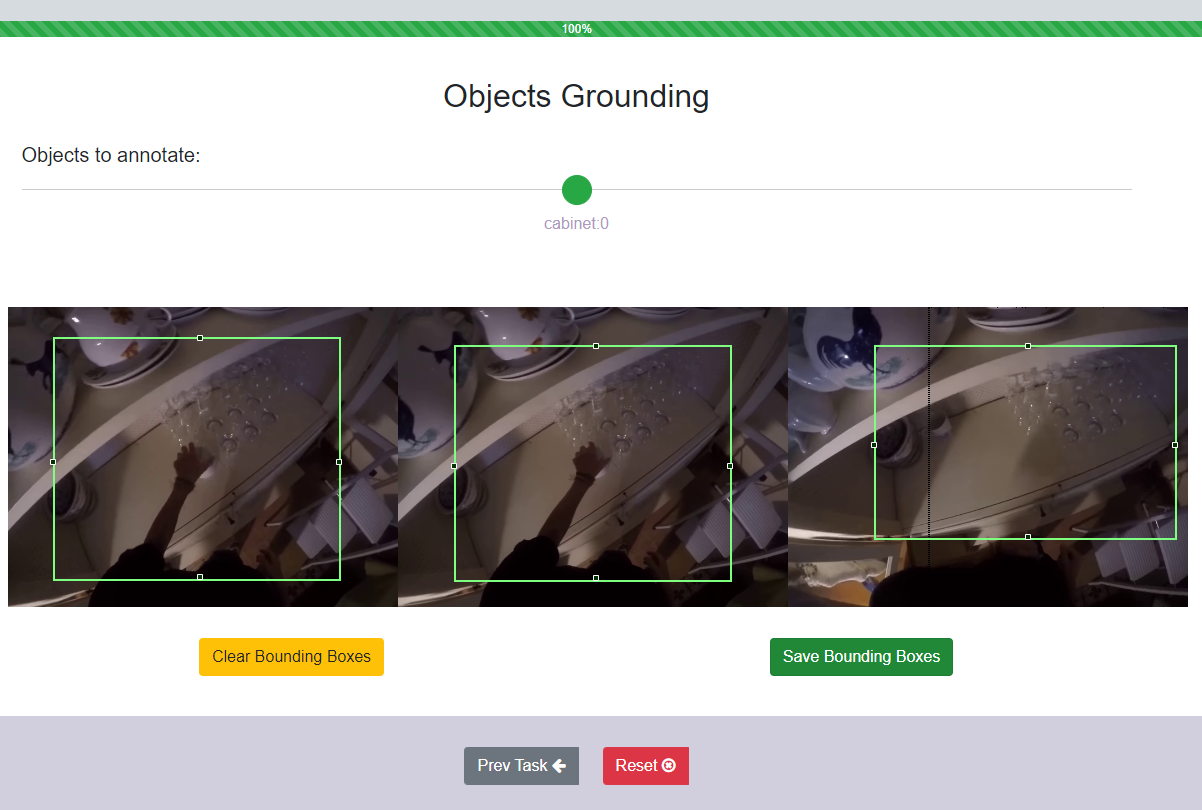}
    \caption{Providing grounding for indirect objects}
    \label{fig:figure5}
\end{subfigure}%
\begin{subfigure}[b]{0.49\textwidth}
    \centering
    \includegraphics[width=\linewidth]{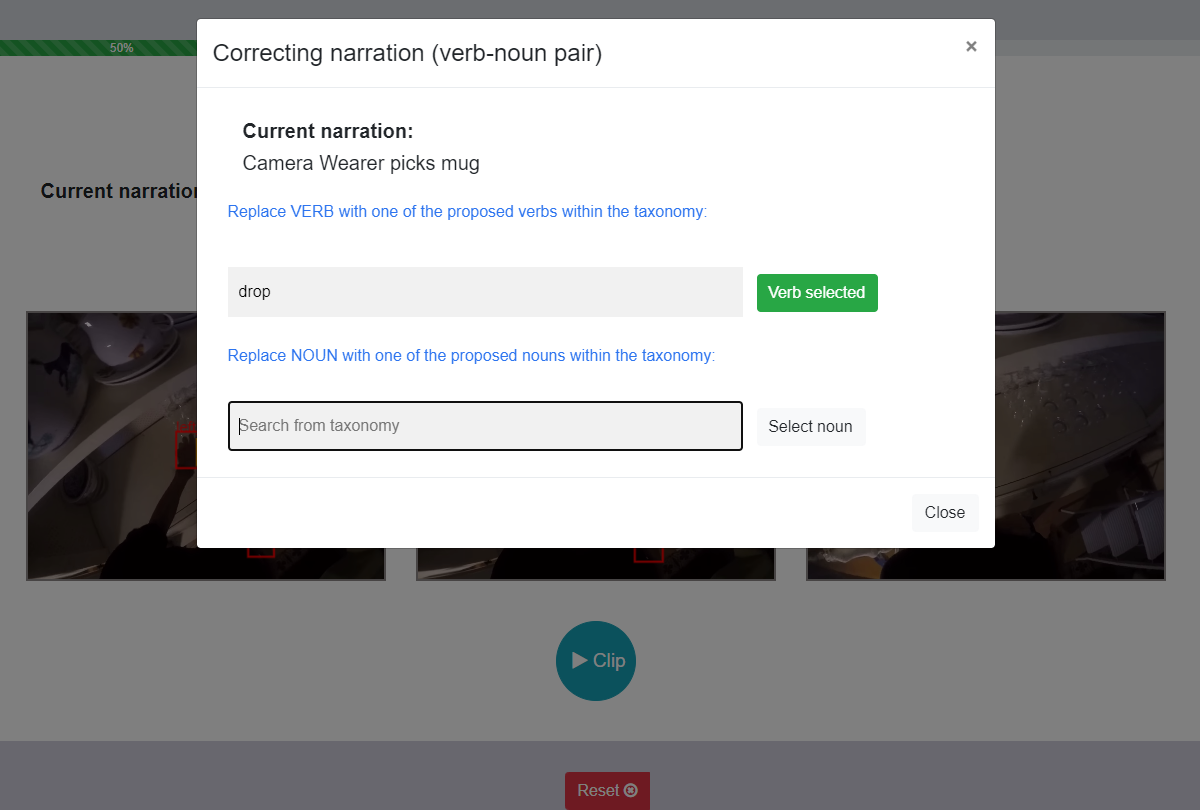}
    \caption{Interface for correcting initial verb-noun pair}
    \label{fig:figure6}
\end{subfigure}

\caption{The procedure which the annotators have to follow in order to provide scene graph annotations. (a) Initial interface providing instructions. (b) A video clip sampled around the PNR frame is shown. (c) The annotator can select among a set of possible relations. (d)~Indirect objects are added by selecting among a list of proposals extracted from narrations or searching from taxonomy. (e) Each indirect object is grounded by the annotator in the three PRE, PNR, and POST frames. (f) In case the provided verb-noun pair is incorrect, the annotators can specify an alternative correct pair.}
\label{fig:screenshots}
\end{figure*}

Figure~\ref{fig:screenshots} reports some screenshots from the annotation procedure. The procedure follows different steps as described in the main paper and illustrated in the figure. An interface providing instructions is initially shown to the annotator (Figure~\ref{fig:screenshots}a). The annotator can hence play a video clip sampled around the PNR frame (Figure~\ref{fig:screenshots}b). They are then prompted to select among a set of possible relations (Figure~\ref{fig:screenshots}c). The annotator can add indirect objects by selecting among a list of proposals extracted from narrations or searching from taxonomy (Figure~\ref{fig:screenshots}d). They will hence ground each indirect object in the three PRE, PNR, and POST frames (Figure~\ref{fig:screenshots}e). If the provided verb-noun pair is incorrect, the annotators can specify an alternative correct pair (Figure~\ref{fig:screenshots}f).

\section{EASG Examples}
\label{sec:dataset-examples}

\begin{figure*}[!t]
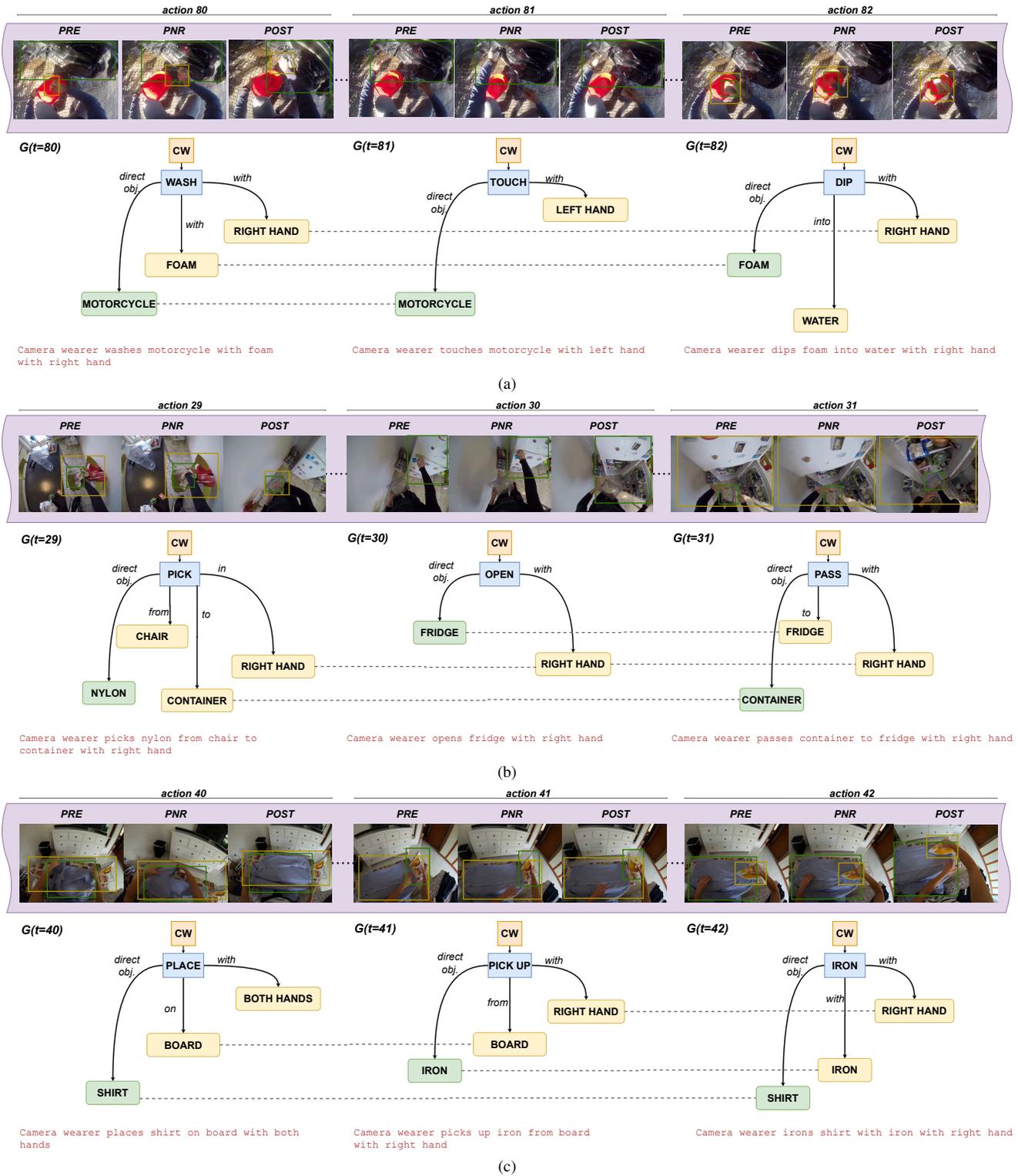
 
    \centering
    \begin{subfigure}{\textwidth}
        \includegraphics[width=\linewidth]{supplementary/EASG_fig1_supp1.drawio.pdf}
        \caption{}
        \label{fig:sub1}
    \end{subfigure}
    \hfill
    \begin{subfigure}{\textwidth}
        \includegraphics[width=\linewidth]{supplementary/EASG_fig1_supp2.drawio.pdf}
        \caption{}
        \label{fig:sub2}
    \end{subfigure}
    \hfill
    \begin{subfigure}{\textwidth}
        \includegraphics[width=\linewidth]{supplementary/EASG_fig1_supp3.drawio.pdf}
        \caption{}
        \label{fig:sub3}
    \end{subfigure}

    \caption{Sample subsequences from the Ego4D-EASG dataset. The temporal nature of Egocentric Action Scene Graphs allows to model long-form relations between the objects in the scene and the camera wearer: the indirect objects may become direct (a, b), and vice versa (b,c).}
    \label{fig:examples}
\end{figure*}

Figure~\ref{fig:examples} provides examples of graph sequences sampled from Ego4D-EASG dataset. The temporal nature of the graphs allows to model long-form relations between the objects in the scene and the camera wearer. For instance, indirect objects may become direct (compare Figure~\ref{fig:examples}a with Figure~\ref{fig:examples}b), and vice versa (compare Figure~\ref{fig:examples}b with Figure~\ref{fig:examples}c).

\section{Verbs, prepositions and object nouns of the EASG dataset}
\label{sec:nodesedges}
Table~\ref{tab:verbs_prepositions_objects} reports the extensive list of all verbs, relations and object nouns.

\begin{table*}[tb!]
\centering
\caption{Extensive list of all verbs, relations, and object nouns in Ego4D-EASG dataset}
\label{tab:verbs_prepositions_objects}
\vspace{-0.5em}
\setlength{\tabcolsep}{5pt}
\resizebox{1\linewidth}{!}{%
\begin{tabular}{lp{16cm}}
\toprule 

  Verbs &  \small{add, adjust, align, apply, arrange, attach, beat, bend, break, bring, bring-out, brush, carry, carry-out, carry-up, carve, change, check, chop, clean, clean-off, clear, climb, close, collect, connect, cover, crumple, cut, cut-off, cut-out, detach, dip, dip in, disconnect, divide, drag, drill, drive, drop, drop-out, drop up, dry, dust, empty, examine, fasten, feel, fetch, fill-up, fit, fix, flap, flip, fold, force, glue, grab, grasp, grip, hammer, hang, hit, hold, hold-up, insert, inspect, iron, join, keep, knead, knit, lay, leave, lift, lift-up, loose, loosen, loosen out, losse, lower, mark, measure, mix, mount, move, move-off, move-up, open, operate, pack, paint, pass, peel, pet, pick, pick-out, pick-up, place, place down, plaster, play, point, position, pour, pour down, pour-in, pour-off, pour-out, press, pull, pull-out, push, push-down, push-in, put, put-away, put-down, put-in, put-off, put-on, put-out, raise, read, release, remove, reposition, rest, return, rinse, roll, rotate, rub, sand, scan, scoop, scoop-out, scrap, scrape, scratch, screw, screw-in, scrub, search, separate, seperate, set, sew, shake, shape, shave, shift, shuffle, slice, slide, smoothen, soak, spin, split, spray, spread, spread out, sprinkle, squeeze, squeeze out, stick, stir, store, straighten, straighten-out, streche, stretch, sweep, swing, swirl, switch, switch-off, take, take-off, take-out, take-up, tap, taste, tear-off, test, throw, throw-away, tie, tight, tighten, tilt, touch, transfer, trim, turn, turn off, turn out, turn over, twist, unfold, unhang, unlock, unplug, unscrew, untangle, untie, untighten, unwrap, uproot, use, wash, water, wear, wet, wipe, wipe-off, withdraw, wrap}\\ \hline
  Relations &  \small{direct object, verb, around, from, in, inside, into, on, onto, out, through, to, towards, under, up, verb, with}  \\ \hline
  Objects &  \small{both hands, left hand, right hand,adapter, apron, art, bag, bar, basket, battery, bed, belt holder, bench, bicycle, bicycle wheel, bike, bike part, bin, board, bobbin, bolt, book, booklet, bookshelf, bottle, bowl, bowls, box, boxer, brake, brake shoe pack, branch, bread, brick, broom, brush, bucket, bulb, bunch, button, cabbage, cabinet, cable, cables, caliper, camera, can, cap, car, car part, carburetor, card, cardboard, carpet, carton, case, casing, cello, cement, chaff, chain, chair, charger, chips, chisel, chopping board, clamp, cleaner, clip, cloth, clothe, clothes, clothing material, compartment, computer, connector, container, control, cooker, cord, cot, counter, countertop, cover, cracker, craft, crumbs, cup, cupboard, cutter, cutting board, debris, derailleur, desk, detergent, dirt, dish, dog, door, dough, dough strip, dough strips, drawer, dress, drill, drill bit, driller, drink, driver, drum, dust, dustbin, dustpan, egg, engine, fabric, fabrics, faucet, fence, file, filter, finger, floor, flour, flower, foam, food, fork, fridge, fuel, furniture, gasket, gauge, gear, generator, glass, glasses, glove, glue, gouge, grass, grater, grease, grinder, ground, guitar, hammer, hand, handle, hanger, heap, hoe, hoes, holder, hose, ice, ice cubes, ice tray, insulator, iron, iron box, ironbox, ironing board, jack, jacket, jar, jug, keg, key, keyboard, knife, knob, knot, ladder, laptop, layer, leaf, leg, lever, lid, lift, light, liquid, lock, machine, manual, marker, mask, mat, matchstick, material, measure, meat, metal, metal board, milk, mirror, mixer, mixture, mop stick, motorbike, motorcycle, mouse, mouth, mower, mug, multimeter, nail, napkin, needle, net, newspaper, note, nozzle, nut, nylon, oil, onion, oven, pack, pad, paddle, paint, paint brush, paintbrush, palette, pan, pants, paper, papers, part, pastry, pedal, peel, peeler, pen, pencil, phone, photo, picture, piece, piece of cloth, pieces, pile, piler, pin, pipe, pizza, plank, plant, planter, plastic wraps, plate, plates, platform, plier, pliers, plug, pocket, pole, polythene, pot, potato, pruner, pruning sheer, pump, purse, rack, rag, rail, railing, rake, refrigerator, rim, ring, rod, rod metal, roller, room, root, rope, ropes, rubber, ruler, sachet, salt, sand, sandpaper, sauce, saucer, saw, scaffold, scale, scarf, scissors, scoop, scooter, scourer, scraper, scrapper, screw, screwdriver, seat, seed, sellotape, serviette, shaft, shear, shears, sheet, shelf, shelve, shirt, short, side, sieve, sink, sink faucet, slab, smartphone, soap, sock, socket, soil, spanner, spatula, spice, sponge, spoon, spring, stack, stairs, stand, steel, stick, stool, stove, strand, string, sugar, switch, table, table cloth, tablet, tag, tank, tap, tape, terminal, thread, tie, timber, timer, tin, tire, tissue, tomato, toolbox, tools, top, torch, towel, toy, train, trash, tray, trey, trimmer, trolley, trouser, trowel, trunk, tub, tube, tyre, umbrella, vacuum cleaner, valve, vase, vice, vine, waist, wall, wallet, wardrobe, washer, water, water hose, weed, wheel, windshield, wipe, wiper, wire, wire cutter, wires, with, wood, wood plank, wooden block, wooden stand, workbench, worktop, wrapper, wrench, yarn, yeast} \\
\bottomrule
\end{tabular}
\label{tab:lists}}
\end{table*}

\section{Prompts for Anticipation and Summarization}
\label{sec:experimental-supp}
%
\newcommand{\cellbreak}{\newline}

\begin{table*}
\centering
\caption{Examples of prompts and outputs for the anticipation task.}
\label{tab:prompts_anticipation}
\scriptsize 
\begin{tabular}{P{4cm}P{5cm}P{4cm}}
\hline
\textbf{System prompt} & \textbf{Input Example} & \textbf{Completion Example} \\ \hline

You are an assistant which models human behaviour very well.
You'll be provided with a sequence of graphs (1..N-1) describing the actions retrieved from a first-person view video.
Your task is to predict the next graph (N). & Example:\cellbreak
Graph 1: Camera wearer - verb - take; take - direct object - flour; take - from - package; take - with - right hand\cellbreak
Graph 2: Camera wearer - verb - add; add - direct object - flour; add - to - bowl; bowl - with - dough; add - with - right hand\cellbreak
Graph 3: Camera wearer - verb - press; press - direct object - dough; press - with - both hands
Graph 4: Camera wearer - verb - move; move - direct object - dough; move - from - bowl; move - to - scale\cellbreak
Graph 5: Camera wearer - verb - move; move - direct object - dough; move - from - bowl; move - to - scale & Prediction:\cellbreak
Graph 6: Camera wearer - verb - remove; remove - direct object - dough; remove - from - scale; remove - to - bowl \\ \hline

You are an assistant which models human behaviour very well.
You'll be provided with a sequence of verb-noun pairs (1..N-1) describing the actions retrieved from a first-person view video.
Your task is to predict the next action (N). & Example:\cellbreak
Action 1: take  flour\cellbreak
Action 2: add flour\cellbreak
Action 3: press dough\cellbreak
Action 4: move dough\cellbreak
Action 5: put dough  & Prediction:\cellbreak
Action 6: remove dough \\ \hline

\end{tabular}
\end{table*}

\begin{table*}
\centering
\caption{Examples of prompts and outputs for the summarization task.}
\label{tab:prompts_summarization}
\scriptsize 
\begin{tabular}{P{4cm}P{5cm}P{4cm}}
\hline
\textbf{System prompt} & \textbf{Input Example} & \textbf{Completion Example} \\ \hline
You are an assistant who can model human behaviour very well.
You'll be provided with a sequence of actions retrieved from
 a first-person view video. Your task is to understand the general activity and describe it
 in one sentence. Please, provide a very general summary and try to avoid listing all the "atomic" activities. & Example:\cellbreak Action 1: Camera wearer pick up hose\cellbreak
Action 2: Camera wearer point hose towards car\cellbreak
Action 3: Camera wearer spray car with water hose\cellbreak
Action 4: Camera wearer wash car\cellbreak
Action 5: Camera wearer raise wiper\cellbreak
Action 6: Camera wearer wash car\cellbreak
Action 7: Camera wearer push down wiper & Summary:\cellbreak Camera wearer is washing and cleaning a car with a water hose and wiper.\\ \hline

You are an assistant which can model human behaviour very well.
You'll be provided with a sequence of verb-noun pairs describing the actions retrieved from
 a first-person view video. Your task is to understand the general activity and describe it
 in one sentence. Please, provide a very general summary and try to avoid listing all the "atomic" activities. & Example:\cellbreak
Action 1: pick up hose\cellbreak
Action 2: point hose\cellbreak
Action 3: spray car\cellbreak
Action 4: wash car\cellbreak
Action 5: raise wiper\cellbreak
Action 6: wash car\cellbreak
Action 7: push down wiper & Summary:\cellbreak Camera wearer is washing and cleaning a car with a water hose and wiper.\\ \hline

\end{tabular}
\end{table*}

All the prompts we are using contain one example of an input sequence and of completion. We provide these examples in order to ensure the correct format of output sequences for the downstream tasks of action anticipation and long-form summarization. Hence, every prompt consists of four parts: 1) Task description; 2) Input example; 3) Output example; 4) Input sequence for which the request is sent. Table~\ref{tab:prompts_anticipation} and table~\ref{tab:prompts_summarization} summarize descriptions, input and output examples for the considered anticipation and summarization tasks respectively.


\section{Qualitative Results for EASG Generation}

\begin{figure*}[htbp]
  \small
    \includegraphics[width=\linewidth]{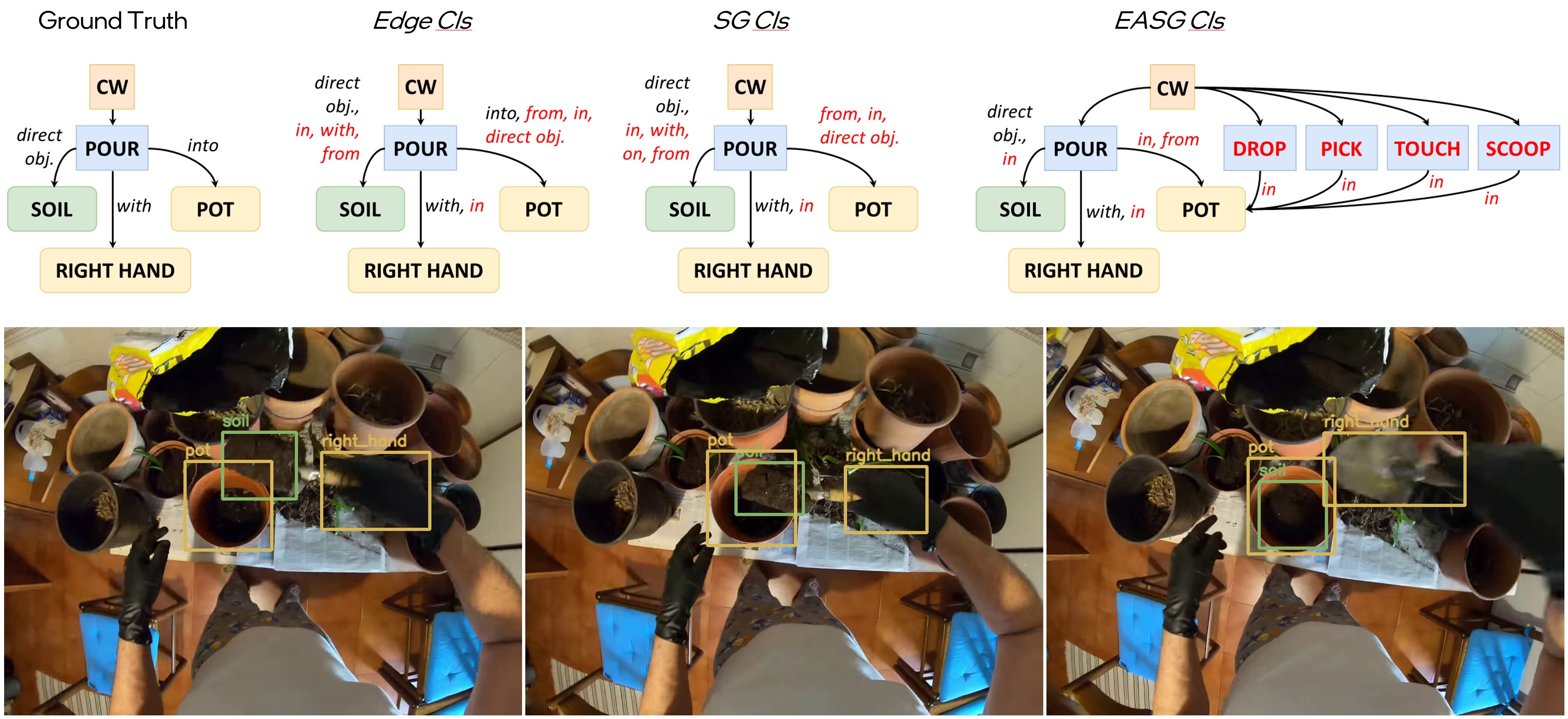}\\ \hspace*{27em}(a) \\[3ex]
    \includegraphics[width=\linewidth]{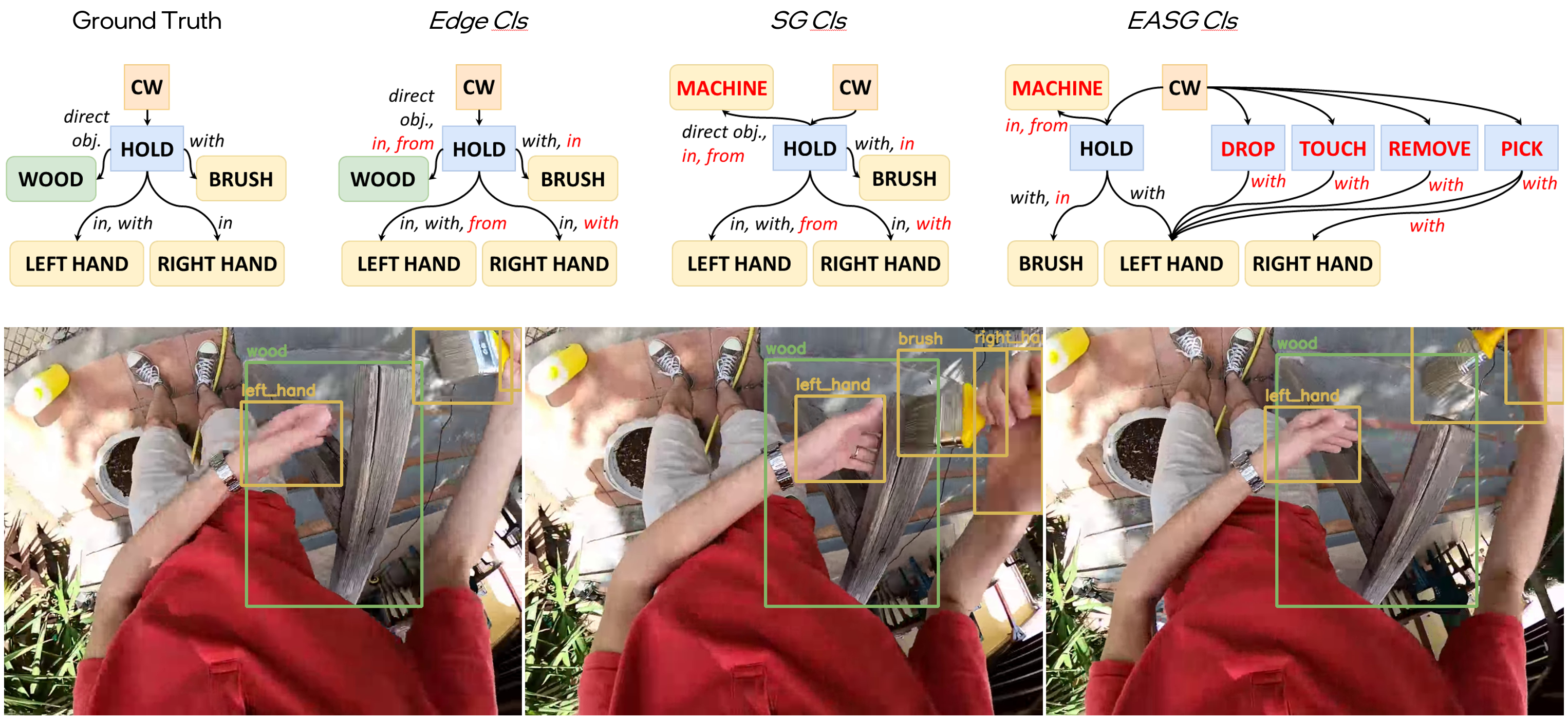}\\ \hspace*{27em}(b)
  \caption{Qualitative results of our baseline model for the three EASG generation tasks (i.e. \textit{Edge Cls}, \textit{SG Cls}, and \textit{EASG Cls}). (a): An example clip of ``Camera wearer pours soil into the pot with right hand.'', (b): An example clip of ``Camera wearer holds wood with the brush in left and right hands.'' Only the top 10 predictions are illustrated in each graph, and texts in red color denote the false positives. We can observe that the generated graphs for the \textit{EASG Cls} task have many false positives when compared to the other two tasks.}
  \label{fig:easg_qual}
\end{figure*}

We provide qualitative results of our baseline model in Figure~\ref{fig:easg_qual}. We draw each graph using the top 10 predictions under the \textit{No Contraint} setup. We can observe that the generated graphs for \textit{EASG Cls} have more false positives than the other two tasks, indicating that action verbs play a significant role in EASG understanding.


\begin{figure*}
    \centering
    \includegraphics[width=\linewidth]{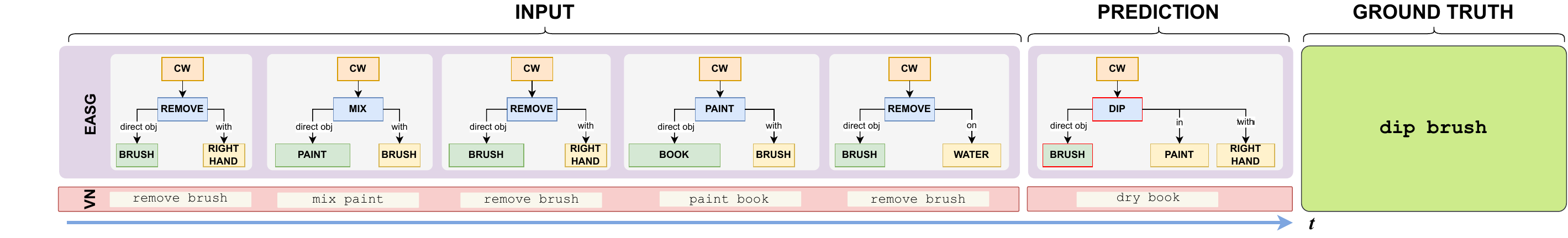}
    \caption{Qualitative example of input sequences and outputs produced using the EASG (top) and verb-noun (bottom) representations for action anticipation. The additional context provided by indirect objects and relations allows the model to predict a more meaningful future action.}
    \label{fig:qualitative_anticipation_2}
\end{figure*}

\section{Qualitative Results for Downstream Tasks}

\begin{figure*}
    \centering
    \includegraphics[width=\linewidth]{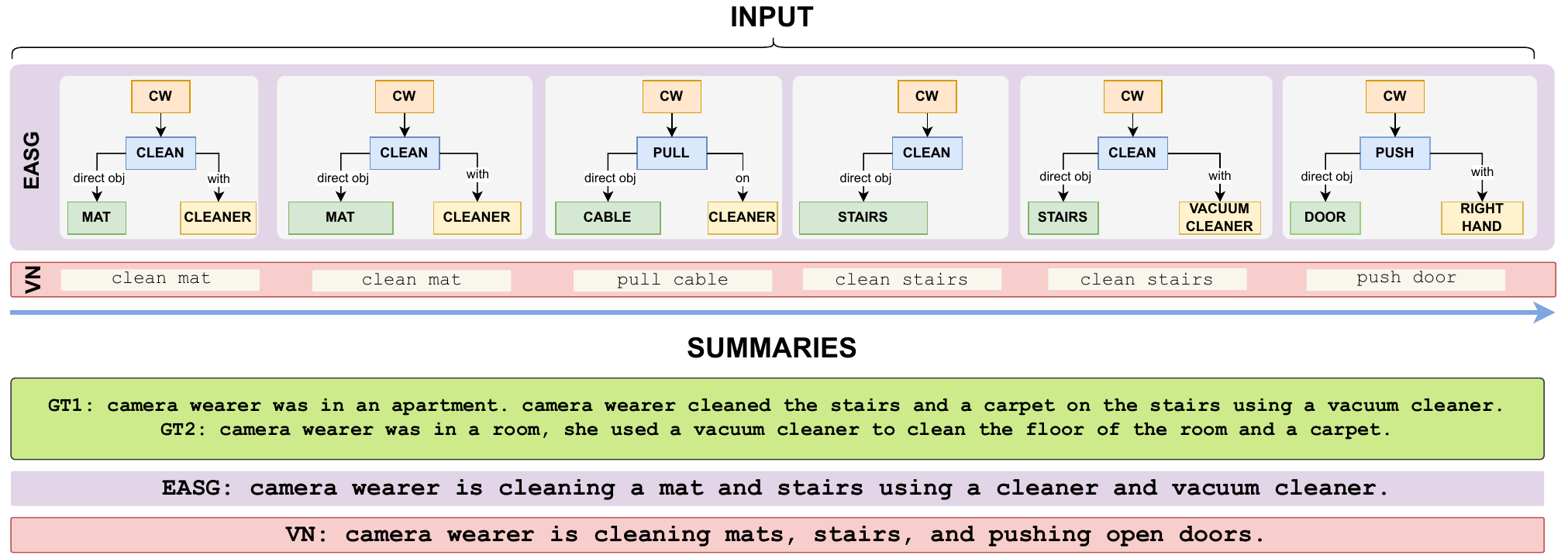}
    \caption{Qualitative example of input sequences and outputs produced using the EASG (top) and verb-noun (bottom) representations for video summarization, along with the reference summaries (in green). Even a single node in EASG (\textit{vacuum cleaner}) may provide an important context for a better understanding of the whole activity.}
    \label{fig:qualitative_summarization}
\end{figure*}

EASGs provide important context for better understanding the whole activity in general. Classic atomic verb-noun form action representations allow to focus on activities performed and human-object interactions, but they often miss important context required for long-form video understanding. In our dataset, there are examples of graph sequences where the important term mentioned in the clip summary does not appear as the active object.
Figure~\ref{fig:qualitative_anticipation_2} and Figure~\ref{fig:qualitative_summarization} show qualitative anticipation and summarization examples respectively.

\end{document}